\RequirePackage{fix-cm}
\documentclass[twocolumn]{svjour3}          
\smartqed  

\usepackage{graphicx}
\usepackage{amsfonts}
\usepackage{mathtools}
\usepackage{hyperref}
\usepackage{amsmath}
\usepackage{epstopdf}

\usepackage{color}
\definecolor{red}{rgb}{0.8,0,0}
\definecolor{green}{rgb}{0.0,0.5,0}
\definecolor{blue}{rgb}{0.00,0.00,0.75}
\definecolor{orange}{rgb}{0.72,0.22,0.06}
\definecolor{purple}{rgb}{0.6,0.0,0.6}
\definecolor{pink}{rgb}{0.58,0.12,0.3}
\begin{document}
\sloppy
\title{Learning Icons Appearance Similarity
}


\titlerunning{Learning Icons Appearance Similarity}        

\author{Manuel Lagunas$^{1}$\qquad Elena Garces$^2$\qquad Diego Gutierrez$^1$
\\
$^1$ Universidad de Zaragoza, I3A\qquad $^2$ Technicolor
}  

\authorrunning{M. Lagunas, E. Garces \& D. Gutierrez} 

\institute{M. Lagunas \at 
			\email{mlagunas@unizar.es}
			Tel.: (0034)976762353
}

\date{}

\maketitle
\begin{abstract}
Selecting an optimal set of icons is a crucial step in the pipeline of visual design to structure and navigate through content. 
However, designing the icons sets is usually a difficult task for which expert knowledge is required. 
In this work, to ease the process of icon set selection to the users, we propose a similarity metric which captures the properties of style and visual identity.
We train a Siamese Neural Network with an on-line dataset of icons organized in visually coherent collections that are used to adaptively sample training data and optimize the training process.
As the dataset contains noise, we further collect human-rated information on the perception of icon's similarity which will be used for evaluating and testing the proposed model.
We present several results and applications based on searches, kernel visualizations and optimized set proposals that can be helpful for designers and non-expert users while exploring large collections of icons. 
\end{abstract}


\section{Introduction}~\label{chap:intro}

Visual communication is one of the most important ways to share and transmit information~\cite{lupton2015,lupton2004}. In the same way as words are used for verbal communication, symbols or icons are the elements used to convey information in a universal and ubiquitous language~\cite{Airey15,horton1994}. 
Icons are key elements to structure visual content and make it more appealing and comprehensible. Thus,
finding the optimal set of icons is a very delicate task usually done by expert designers which involves
semantic, aesthetic, and usability criteria. Recent works aim at automatizing this task and make it more accessible to the general public~\cite{bates2002,Setlur2005,Setlur2014,mohler2015screen}, either by providing a unified icon representation and rules, such as Google Materials\footnote{https://material.google.com/},
or with online datasets such as The Noun Project\footnote{https://thenounproject.com/}
with more than one million elements. 
While these datasets are undoubtedly useful, they can be hard to explore due to their magnitude. 

\begin{figure}[tb]
	\centering
	\includegraphics[width=0.8\linewidth]{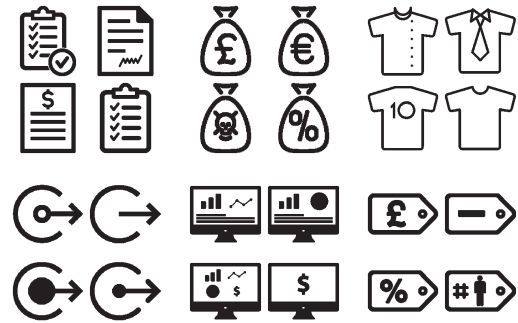}
	\caption[Example of six different collections of the dataset.]{Example of six different collections of the dataset. Style and visual identity are preserved for each collection. From left to right, we see the collections labeled as: \emph{notebook}, \emph{bags}, \emph{t-shirt}, \emph{circle-arrow}, \emph{monitor}, and \emph{label}.}
	\label{fig:introClases}
\end{figure}

The following properties are desirable for an icon set to be effective: first, being appropriate for the meaning -usually, the icon's designer provide semantic labels. Second, being visually appealing by means of a coherent \textit{style} and a carefully defined \textit{visual identity}~\cite{Setlur2014}.
As seen in the literature~\cite{Airey15}~\cite{barnard2013}, we define style as the set of pictorial features in the icons such as stroke, fill, or curvature; and visual identity as the property that makes a set of icons visually identifiable and unique, it is a higher-level property usually linked to the shape of the object.
Previous works have studied style in fonts~\cite{odonovan2014}, clip art~\cite{GarcesSIG2014}, or infographics~\cite{Saleh2015}. Although the definition of style for these domains shares certain properties with icons
style, e.g. strokes, fills, or corner smoothness; icons have additional
characteristics that make them unique and visually identifiable, and these are not taken into account in the existing metrics.
For example, in Figure~\ref{fig:introClases}, the collections \textit{notebooks} and \textit{bags} have a different visual identity while their pictorial style can be considered similar. Note that each icon also has a unique semantic meaning independent of the collection's name.

On the other hand, the problem of choosing optimal icon sets is a recent topic of research. Previous works~\cite{demiralp2014learning}~\cite{laursen2016} have proposed perceptual kernels for predefined icon sets based on crowd-sourced data. These techniques learn directly a similarity matrix (or kernel) strictly for the icon selection. As they do not find a new low-level feature space for each icon, these techniques are not able to generalize outside the initial sample space of ten or twenty icons.

In this work, we present a 
learning-based 
similarity metric that captures the properties of style and visual identity for iconography. 
Our main contributions are:
\begin{itemize}
	\item We present an icon dataset labeled by designers where each collection shares a coherent style and visual identity.
	\item We learn icons' appearance similarity using a	Siamese Neural Network with a triplet loss function and adaptive sampling trained from our weakly-labeled dataset and evaluated with human ratings. 
	\item We propose several applications including search by similarity and 
	a method to create icon sets optimized for style and visual identity in order to help users on user-interface design tasks.
	\item We collect annotated ratings on the perception of appearance similarity for iconography.
\end{itemize}
We greedily gather an icon dataset from the Noun Project online database. Since the semantics of each icon is highly attached to the application, we assume that each icon is labeled with a keyword that represents its concept properly. The icons in this dataset are organized in collections, which share a style and have a particular visual identity (see Figure~\ref{fig:introClases}). 
As previous methods do not fully consider the pictorial properties of icons, we use the collected dataset to train a new Siamese Neuronal Network by adaptively sampling meaningful triplets of relative comparisons. 
However, as the labeling of the collections is very noisy, 
-there is no unified and homogeneous label set that we can completely trust- we need to gather new reliable data for testing the model.  
We numerically evaluate the performance of our distance metric on this test data, and compare its performance to existing similarity metrics. 
Finally, we propose an application to optimize icon sets for the properties of style and visual identity that can be used as a tool to help users while designing graphical interfaces. To validate the method we launch a crowd-sourced survey to a group of 25 human-raters with experience in Computer Graphics or Graphic Design. Users reported that our method returns a set of icons sharing a representative appearance 75.25\% of the times, while random icon sets share a representative appearance 29\% of the times.


\section{Related Work}\label{chap:related}

\paragraph{Icon Design}
Previous works have focused on generating semantically relevant icons to improve visualizations~\cite{Setlur2005,Setlur2014}. In particular, Setlur and Mackinlay~\cite{Setlur2014} develop a method for mapping categorical data to icons. They found out that users prefer stylistically similar icons within a set, as opposed to automatic sets that might differ in look-and-feel. Lewis et al.~\cite{Lewis2004} studied how the perception of icons is affected by spatial layouts, and present a shape grammar to generate visually distinctive icons. Our work is inspired by these, although we propose a deep learning-based method to measure style and visual identity between icons.

More recently, the work of Liu et al.~\cite{Liu2016} proposes a semi-automatic method to create icons from images according to a given style, while the work of Bernstein and Li~\cite{bernstein2015lillicon} describes a technique to make icons scale independent. 
Our technique is complementary to those as can be used as an evaluation metric.

\paragraph{Style Similarity} 
Style similarity metrics have been recently proposed for fonts~\cite{odonovan2014},
infographics~\cite{Saleh2015}, 3D models~\cite{Lun2015,Liu2015}, or interior designs~\cite{bell2015learning}. 
Closer to our goal, the work of Garces et al.~\cite{GarcesSIG2014} uses a hand-made feature vector to measure style similarity for clip art. However, since the feature descriptors were manually selected for that particular task, and do not account for high-level properties, their distance metric does not generalize to our data, as we will show later. In a follow-up work, Garces et al.~\cite{garces2016style} find that shape is a property that people take into account when comparing clip arts, however, it is not measured in their existing style metric for clip art.
On the contrary, we automatically learn a distance metric that measures both style and visual identity using a deep Siamese Neural Network trained from scratch.

\paragraph{Shape Similarity}
To measure shape similarity is a long-standing problem in computer graphics with many different approaches trying to solve it. Bober~\cite{Bober01} shows how to represent and match shape representations under the MPEG-7 standard~\cite{Sikora01}. Osada et al.\cite{Osada2002} propose several silhouette-based descriptors that can be used for 2D and 3D shape retrieval.
Other shape descriptors have been proposed, including Hu-moments~\cite{hu1962visual},
shape context~\cite{belongie2002shape}, 
the use of Zernike moments~\cite{khotanzad1990invariant},
pyramid of arclength descriptors~\cite{Kwan2016}, or Fourier descriptors~\cite{zhang2002shape}. Kleiman et al.~\cite{kleiman2015} focused on 3D shape similarity, using part-based models, while other works compare shapes using single closed contours~\cite{latecki2000shape,bai2010learning}. In contrast, our method does not need to explicitly model the geometrical properties of the given image and implicitly considers additional properties such as image abstraction and complexity that are recognized while training the Siamese Neural Network.

\paragraph{Kernel Learning}
In contrast to the previous works that rely on a feature-based representation of the data, kernel methods aim to obtain directly the similarity matrix for a fixed set of objects, thus such approaches do not generalize to objects outside the chosen set~\cite{gramazio2017}~\cite{shugrina2017}. The work of Laursen et al.~\cite{laursen2016} proposes an embedding of a small fixed set of icons optimized for comprehensibility and identifiability properties. Demiralp et al.~\cite{demiralp2014learning} re-order icon sets to maximize perceptual discriminability. Closer to ours, non-linear content-based retrieval methods use similarity metrics tied to the context of their particular problem~\cite{Elnaqa04,Wu13,Doulamis04,Xia14}. Unlike our work, kernel methods learn directly the distance over the given set of objects relying on user judgments. While we propose a general metric trained on a large set of icons and based on deep image representations learned by the Convolutional Neural Networks. Our metrics are valid for any candidate, even outside the sample space.


\section{Problem Definition}

Our main goal is to obtain a metric to measure style similarity and visual identity between icons. As mentioned in Section~\ref{chap:intro}, an icon can be defined by its pictorial properties like outline stroke, fill or curvature~\cite{bernstein2015lillicon}, features that conform the pictorial style of the icon. 
In addition, 
a set of icons is also characterized by a particular visual identity~\cite{barnard2013}~\cite{Airey15}, i.e. one or more properties that make it unique and visually identifiable. Commonly, these properties relate to a particular shape or a motif, which repeats between icons of the same collection e.g. a silhouette circle, a notebook-like shape, an arrow, etc. (see Figure~\ref{fig:introClases}).

\begin{figure}
	\centering
	\includegraphics[width=0.85\linewidth]{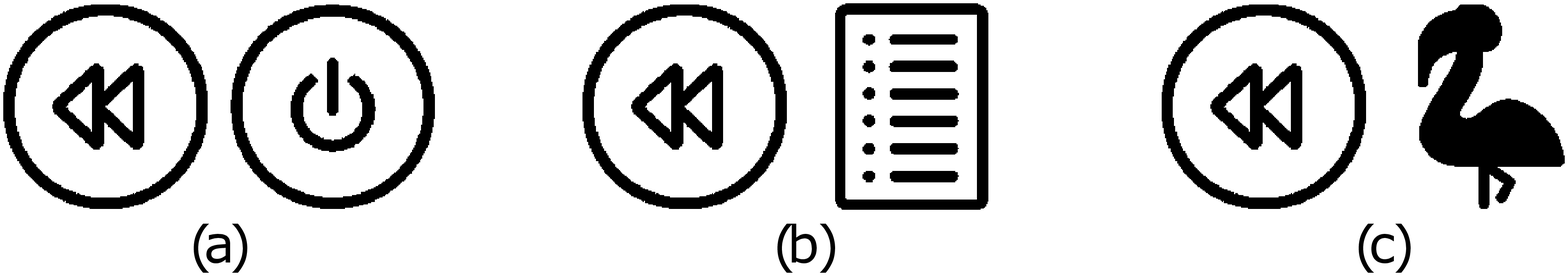}
	\caption[Examples of similarity between icons.]{Examples of similarity between icons. (a) Icons with similar style and visual identity. Note that both icons have rounded shapes and medium-thick lines. (b) Icons with similar style yet different identity, one has rounded shape while the other is a rectangle. (c) Icons whose style is different and they also have different identities. 
	}
	\label{fig:pbmStatement}
\end{figure}

Finding clusters of perceptually different icon sets 
is really impractical given the subtle differences between them.
Instead, as seen in previous work~\cite{GarcesSIG2014,odonovan2014,Liu2015,Saleh2015}, it is more intuitive to find a continuous metric space where the distances between the icons correspond to distances in the perceived similarity. 
Given that previous definitions of style use hand-crafted features for other domains that do not apply for icons, we aim to find a new similarity metric $\mathcal{D}$ that measures differences in style and differences in visual identity:
\begin{equation}\label{eq:distance}
\mathcal{D}(i, j) = \mathcal{D}_s(i,j) + \mathcal{D}_v(i,j)
\end{equation}
where  $(i, j)$ is a pair of icons,  the function $\mathcal{D}_s(i,j) \in \mathbb{R}^+$ measures style similarity, and the function $\mathcal{D}_v(i,j) \in \mathbb{R}^+$ measures visual identity. 
For icons with similar style and visual identity, $\mathcal{D}$ should return small values, i.e. $\mathcal{D}_s\simeq0$ and $\mathcal{D}_v\simeq0$ (Figure~\ref{fig:pbmStatement},~a). For icons with similar style but with different identity, 
$\mathcal{D} = \mathcal{D}_v$ (Figure~\ref{fig:pbmStatement},~b). Finally, for icons where both properties are very different, the similarity function will also have a high value; $\mathcal{D} \gg 0$ with $\mathcal{D}_v \gg 0$ and $\mathcal{D}_s \gg 0$ (Figure~\ref{fig:pbmStatement},~c). 

\subsection{Overview}
\begin{figure*}[!htb]
	\centering
	\includegraphics[width=0.95\linewidth]{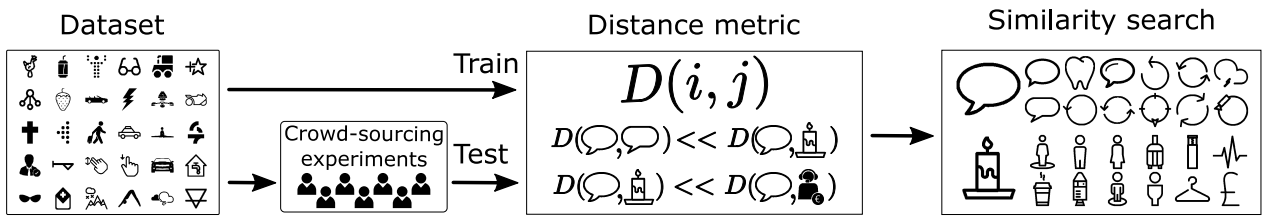}
	\caption{Overview of our work: The leftmost part shows the data gathering process. First, we collect a dataset of icons and use it to train the similarity metric. Since the dataset contains icons labeled by the designers, we cannot completely trust their annotations and might find spurious data or noise. Due to that, we use part of the data gathered to launch crowdsourcing experiments in Amazon Mechanical Turk and obtain curated test data that we use to compare the trained models. Once the data is collected, we train a Siamese Neural Network (SNN) that works as our distance metric, returning small values for icons that share style and visual identity while returning large values for icons that do not share those properties. With the trained model we are also able to compare icons distances and perform similarity searches by returning the icons with the minimum distance to a reference in the learned Euclidean space.}
	\label{fig:overview}
\end{figure*}

An overview of the method can be seen in Figure~\ref{fig:overview}. Our main goal is to obtain a similarity metric $\mathcal{D}(i,j)$ where $i,\;j$ are a pair of icons. 
To train the similarity metric, we use a dataset which is annotated by icon designers. Since there is no unified way of labeling, we cannot completely trust the annotations and we might find noise in some of its classes. 
This kind of datasets are called \textit{weakly labeled} and additional efforts are required to work with them. 
In our case, part of the dataset is used to launch crowd-sourcing surveys and gather human-ratings that will allow us to test and compare the proposed models (Section~\ref{chap:data}). The other part of the data will serve to train a Siamese Neural Network (SNN) to work as the similarity metric (Section~\ref{chap:method}). The SNN maps the input icons into a new Euclidean feature space where they can be compared. The new mapping of the icons can be further used to propose different applications like searches by similarity, or propose icon sets optimized for the properties of style and visual identity.

The concept of weakly-labeled data might resemble weakly-supervised learning~\cite{Crandall06,Torresani2014,Voulodimos18}. However, in weakly-supervised learning we have a constrained amount of annotated data, on the other hand, weakly-labeled data has no annotations but we know some meta-information about each sample. Moreover, in weakly-labeled data, we do not have any constraints on the amount of data used during training.


\section{Collecting Data}\label{chap:data}

We obtain our icon dataset from the \textit{Noun Project} website, which contains thousands of black and white icons uploaded by graphic designers. Using the provided API we greedily downloaded a total of 26027 different icons, grouped in 1212 collections or classes each one sharing a label decided by the author (see Figure~\ref{fig:introClases} for a few examples). 
Each icon belongs to just one class and most of the icons per class share similar style and visual identity properties. 
As a first step, 
by means of stratified sampling, we split the dataset into three subsets: training (70\%), validation (10\%), and test (20\%).
We consider each class as the strata, then, we randomly select elements from each class proportionally (according to the given percentages) to sample the train, validation and test subsets. All the elements in each class are sampled and the subsets are mutually exclusive, meaning that each element is sampled only once and for one of the subsets.
However, the labels provided by the designers are not disjoint and we might find different labels with the same style and identity and one label with different styles or identity. This kind of \emph{weakly-labeled}~\cite{SimoSerra2016} data may yield
problems like not detecting if the model has overfitting or not allowing a fair comparison with other architectures at testing time. Thus, 
further data collection and adjustments are needed to take full advantage of the dataset.

\paragraph{Collecting Curated Data}
We collect valid data on the perception of icon's similarity that will be used to test the proposed models and select the best one.
We use \textit{Amazon Mechanical Turk (MTurk)} to launch the experiments. Similar to previous works~\cite{GarcesSIG2014}~\cite{bell2015learning}~\cite{Liu2015}, we gathered data in the form of \textit{relative comparisons}, since they are more robust and easier for human raters than Likert ratings~\cite{demiralp2014learning}~\cite{Rubinstein10}.
The structure of each test, or HIT, consisted of: first, a clear description of the task that human raters had to perform, then, a training phase where we show a small set of four manually picked relative comparisons displaying guidance messages if the user fails answering correctly. The last part corresponds to the test phase, where the rater has to answer a total of 60 relative comparisons where seven questions belong to a manually selected control set with an obvious answer. The duration of each HIT was approximately seven minutes, and we paid an average of \$0.15 per HIT.

We rejected all human raters that had more than one error (out of seven) in the control questions. In the end, we launched 6000 relative comparisons tests each of them answered by ten users, 962 HITs were approved and 38 rejected. To create the relative comparisons for each question, we randomly selected one icon per class from three different random classes. We allowed participants to do as many HITs as they wanted without repetition.
A total of 213 users took part in the survey, 43\% female. Among raters, 5.95\% claimed some professional experience in user interface and interaction design, while 6.43\% have had some professional experience with graphic design. 


\section{Modeling Visual Appearance of Icons}\label{chap:method}

Existing style similarity metrics~\cite{GarcesSIG2014,Saleh2015} use a handcrafted feature space only suitable for their respective domains, where only local style features are taken into account. On the contrary, besides style, our metric should measure also visual identity, which is usually a higher-level property related to the shape of the icon. 
Image similarity has been measured with existing deep models, such as VGG19~\cite{Simonyan14}, pre-trained on natural images; and fine-tuning these networks has worked well for tasks such as interior design similarity~\cite{bell2015learning}. 
However, we would need a huge amount of training data to improve the performance of any existing network, and given that our domain is much simpler than pictures of natural images, we choose to train a new network with our data. To make sense of the difference, the widely used network VGG19 has 144M of parameters, while our network has 47M parameters.

\begin{figure}[tb]
	\centering
	\includegraphics[width=1\linewidth]{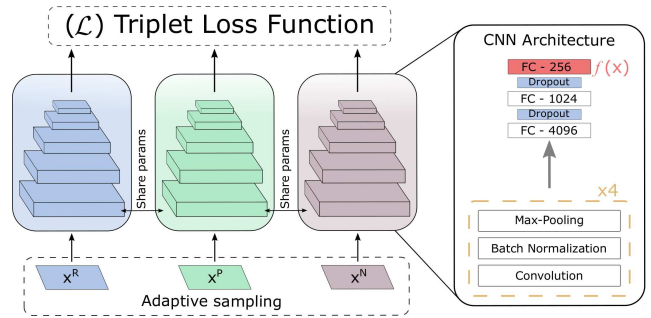}
	\caption{Architecture proposed to measure icons similarity. The Siamese Network has three inputs: Reference ($x^R$), Positive ($x^P$) and Negative ($x^N$); and three Convolutional Neural Networks (CNN) to obtain its embeddings ($f(x)$). With these three embeddings, we can compute the error of the network ($\mathcal{L}$) using the triplet loss function described in Equation~\ref{eq:loss}. The CNNs share the same structure and parameters. Each of them has four convolutional layers, that are followed by a batch-normalization layer and a max-pooling layer. The last pooling layer is connected to a linear classifier with three fully connected layers (FC). First FC has 4096 features, second one has 1024, while the last FC has only 256, furthermore, last FC of each CNN corresponds to the embedding $f(x)$ of the input triplet $[x^R, x^P, x^N]$. Between the FC layers there are dropouts with regularization rate of 30\%.}
	\label{fig:styleArch}
\end{figure}

We use a Siamese Neural Network~\cite{Bromley94}~\cite{Parkhi15}~\cite{Schroff2015} consisting in three identical Convolutional Neural Networks (CNN) that share their parameters. This kind of architecture is really powerful for learning a new Euclidean space~\cite{Schroff2015}~\cite{Parkhi15}~\cite{Fried17} where objects can be compared~\cite{Bromley94}~\cite{Yin15}. Since the icons inside a collection in the dataset share the properties of style and visual identity, the SNN can be trained to map together the icons that share these properties while it separates icons with different style and visual identity.
Each CNN has four convolutional layers that are followed by a batch normalization~\cite{ioffe2015} layer and a max-pooling layer. The last pooling layer is connected to the linear classifier. The linear classifier contains three fully-connected layers where the first two have 4096 and 1024 features respectively. The last layer represents the final embedding $f(x)$ of the image $x$ into the new feature space $\mathbb R^d$, where the value of $d$ has been empirically set to 256.
We also included two dropout~\cite{Srivastava2014} layers between the fully-connected ones with a dropout regularization rate of 30\%. An example of the architecture we described is shown in the Figure~\ref{fig:styleArch}, right.
This architecture is trained using triplets of images: a reference $x^R$, a positive $x^P$ (icon with similar properties to the reference), and a negative $x^N$ (icon with different properties to the reference). To train the network we design a specific loss function which is explained below.

\subsection{The Loss Function}~\label{sec:loss}

Let's consider the output of the last fully-connected layer of the Convolutional Neural Network as an embedding $f(x) \in \mathbb R^d$ with input $x$. The embedding represents $x$ in a new $d$-dimensional Euclidean space. Since we have a Siamese Neural Network formed by three CNNs that are identical with three inputs $[x^R, x^P, x^N]$, we get three embeddings as the output $[f(x^R), f(x^P), f(x^N)]$ where $f(x^R)$ corresponds to the embedding of a reference input while $f(x^P)$ is the embedding of an input of the same class as the reference and $f(x^N)$ is the input of an image that does not belong to the same class as the reference. We want to ensure that a reference icon $x^R$ is closer to every icon of the same perceptual similarity (style and visual identity) $x^P$, than to the rest of icons with different image properties $x^N$. Thus the triplet loss function $\mathcal{L}$ (Equation~\ref{eq:loss}) has to ensure that the distance in the $d$-dimensional Euclidean space between the reference and the positive icon is minimum while it is large between the reference and the negative icon~\cite{Schroff2015,Parkhi15}. 
\begin{equation}
\mathcal L = \sum_{i=1}^{M}\Bigg[ \lVert f(x_i^R)-f(x_i^P)\lVert^2_2 - \lVert f(x_i^R)- f(x_i^N)\lVert^2_2 + \alpha \Bigg]_+
\label{eq:loss}
\end{equation}
Here $M$ is the training set of triplets and $\alpha$ is a margin enforced between negative and positive pairs which was empirically set to 0.2. The value $\alpha$ prevents the function from evaluating to zero in cases where the distance between the reference and the negative sample is larger than the reference and the positive sample, thus letting it find larger margins while training.

\subsubsection{Adaptive Sampling}\label{sec:sampling}

\begin{figure}[!htb]
	\centering
	\includegraphics[width=0.9\linewidth]{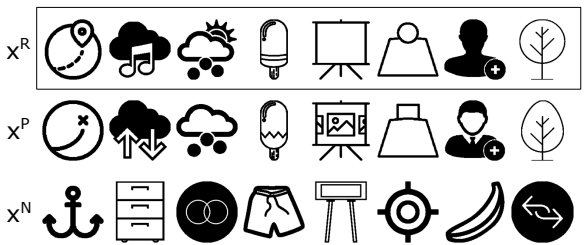}
	\caption[Examples of the triplets sampled during training.]{Examples of the triplets sampled during training. The variables $x^R$, $x^P$ and $x^N$ refers to the reference, positive and negative icon respectively. The positive icon and the reference are selected from the same class and they have the larger Euclidean distance among the icons inside that class. The negative icon has the shorter Euclidean distance to the reference among the icons within a different randomly selected class.}
	\label{fig:tripletSampling}
\end{figure}

If we would like to create all the possible triplets from the, approximately, 18200 icons in the training set we would have ${\binom {18200} {3} \simeq 6.027\cdot 10^{12}}$ possible combinations, an unmanageable number using a standard desktop configuration. Furthermore, most of the generated triplets would easily satisfy the constraints of the loss function and not contribute to the training process at all, thus slowing it. For this reason, following the approach of Schroff et al.~\cite{Schroff2015}, we generate the triplets on the fly during the training process, selecting the ones that are active and help in the convergence. We generate triplets that violate the most the constraints imposed by the loss function. To do so, we randomly select one icon from the training set as the \textit{reference}, then, we select the \textit{positive} sample as the icon from the same class with the maximum distance in the Euclidean space to the reference: $\text{argmax}_{x^P_i}||f(x_i^R)-f(x_i^P)||^2_2$. To obtain the negative icon, we randomly
pick a different class and select the icon that has the minimum distance to the reference: $\text{argmin}_{x^N_i}||f(x_i^R)-f(x_i^N)||^2_2$. We repeat this approach until a considerable number of triplets without repetition has been obtained. This process is applied before every epoch and it requires to compute the embedding for every icon at each iteration. In the first iteration, embeddings are directly obtained from the network whose parameters have been set using Xavier's initialization~\cite{Xavier10}. Although it increases the training time, it also ensures that all input triplets are meaningful for the training. Figure~\ref{fig:tripletSampling} shows an example of the triplets sampled during training in the first iteration.

\subsection{Train the Models}\label{sec:train}
We use ADAM optimization~\cite{Kingma14} and the triplet sampling explained in Section~\ref{sec:sampling}. The mini-batch had a size of sixteen images and to update the parameters of the network we use standard back-propagation~\cite{LeCun1989,Goodfellow2016}.
At training time, we perform two sequential operations with each image before feeding it to the network: first, \textit{data augmentation} (randomly rotating or flipping the image) and second, \textit{random crops}. 
For the crops, we randomly perform a crop of size 180x180 aligned to the corners in the original image, with size 200x200. We started the training with a learning rate of $10^{-4}$ that was reduced every 60 epochs by a factor of ten to let the model converge. To create the validation set we also use the adaptive sampling, moreover, each image is scaled to 180$\times$180 instead of cropped and no data augmentation is applied. We need around two days and 140 epochs to train the model.


\section{Model Testing }

We evaluate the performance of the models by comparing their \textit{precision} and \textit{perplexity} on the gathered data from the MTurk HITs. At testing time, no data augmentation is applied and the inputs are directly scaled to $180\times180$ without cropping.
First, we obtain the embedding for the three inputs of the triplet $[f(x^R), f(x^P), f(x^N)]$, since they are in a $256$-dimensional Euclidean space, we can calculate the Euclidean distance of each icon with respect to the reference $\mathcal{D}(x^R,x^P)$ and $\mathcal{D}(x^R,x^N)$. Actually, if we want to obtain the probability of choosing the icon $x^P$ over $x^N$, what we are aiming to obtain is a function of similarity instead of a distance, thus we define the similarity between two icons $s(x^R, x^P)$ as:
\begin{align}
s(x^R, x^P)= \frac{1}{1+\mathcal{D}(x^R,x^P)}
\end{align}

when the positive $x^P$ and reference $x^R$ icon are completely similar $\mathcal{D}(x^R,x^P)=0$, their similarity is $s(x^R,x^P)=1$. In the opposite case, if the pair of icons is completely dissimilar:
$s(x^R,x^P)=0$. Knowing that $\mathcal{D}(x^R,x^P)$ cannot be infinity, we can define the probability of choosing the icon $x^P$ against $x^N$ as:

\begin{align}
\mathbb P(x^P) = \frac{ s(x^R,x^P)}{ s(x^R,x^P) +  s(x^R,x^N)}\label{eq:P(P)}
\end{align}
We can obtain $\mathbb P(x^N)$ similarly.
Then, we compute precision and perplexity in two ways: assuming the correct answer relies on each turker opinion separately (\textit{raw}) or assuming the majority opinion is the correct one (\textit{majority}). We also compare our results with two baselines previously calculated: the Humans and the Oracle precision. To compute Humans baseline, we count the rater's opinion and compare it to the majority. For the Oracle baseline, we count the opinion of the majority on each relative comparison, being the precision always one.

The precision $\mathcal P$ tells us the percentage of icons that the model has predicted correctly according to our two criteria (\textit{raw} and \textit{majority}).
The precision value is computed as: 
\begin{align}
\mathcal P = \frac{\text{ Icons correctly predicted}}{\text{Number of total relative comparisons}}
\end{align}
The perplexity $\mathcal Q$ is often used for measuring the usefulness of a model when predicting a sample. Its value is 1 when the model makes perfect predictions on every sample, while its value is 2 when the output is 0.5 for every sample, meaning total uncertainty. We define the perplexity of our model as
\begin{align}
\mathcal Q=2^{\big(-\frac{1}{M}\sum_{i=1}^{M}\log_2 \mathbb P(x^P_i)\big) }
\end{align}
To know which one is the positive sample $x^P$ in the relative comparison we rely on raw and majority criteria as for the precision. The value $\mathbb P(x^P)$ will be the probability given by the model using Equation~\ref{eq:P(P)}, $M$ corresponds to the number of triplets we use for testing.

\subsection{Other Architectures}\label{sec:test}
\begin{figure}[!htb]
	\centering
	\includegraphics[width=0.9\linewidth]{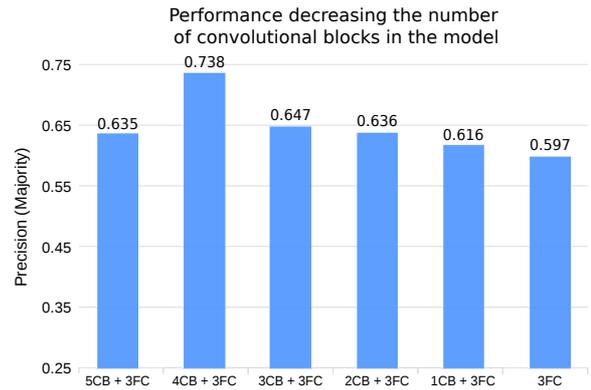}
	\caption[Model performance while varying the number of layers]{Model performance while varying the number of layers. The vertical axis shows the majority precision obtained while the horizontal axis shows the model description. In the models description, \textit{CB} refers to the convolutional Blocks and \textit{FC} to the Fully Connected layers. We can observe how the best model has four convolutional Blocks achieving nearly 74\% majority precision. The models with less number of layers and parameters are not able to reach that performance. Also, the model with five convolutional blocks seems to overfit getting similar performance to the model with just two convolutional blocks.}
	\label{fig:perfComp}
\end{figure}

We followed an incremental approach while designing the Siamese Neural Network. We tested out how the number of convolutional blocks (CB) affects model performance while keeping the same training parameters and same layers in each block (Convolution + Batch norm. + Pooling). Figure~\ref{fig:perfComp} shows how model performance varies, achieving best results with 4 convolutional blocks.

Once we know that the best accuracy is obtained with four convolutional blocks, we explore the performance varying the layers inside each block and the number of Fully Connected layers. Table~\ref{tab:modelsResults} shows the precision and perplexity of the architectures described below. All the included architectures have four convolutional blocks.
\textit{Model-A} has max-pooling between the convolutions and two fully-connected (FC) layers. It has one of the worst results since it does not include layers to avoid overfitting or improve performance with non-linearities. \textit{Model-B} includes max-pooling between convolutions and dropout between the two FC layers. The architecture is similar to Model-A and its result is the worst in terms of both, precision and perplexity. 
\textit{Model-C} includes only max-pooling between convolutions and has three fully-connected layers with dropout between them. The new FC layer does not improve the performance of this model and its results remain lower in comparison to Model-C. Finally, \textit{Model-D} includes max-pooling, batch-normalization and ReLUs between convolutions and it also has dropout between the three FC layers yet it does not improve the performance of Model-D. 

\subsection{Comparing Previous Works}\label{sec:compPW}

In Table~\ref{tab:modelsResults} we also compare our best model with a well-known pretrained architecture VGG19~\cite{Simonyan14} and a hand-crafted feature vector for clip art style~\cite{GarcesSIG2014}. VGG19 model is able to achieve 63\% of precision yet it was not designed to find a space where icons can be compared by similarity and its results are worse than most of the trained architectures. Also, the time needed to get the feature vector of an image is nearly two orders of magnitude higher than with our model, that just needs $9*10^{-4}$ seconds. The method of Garces et al. achieves worse accuracy than VGG19 and our model since the hand-crafted feature space was designed to measure style similarity in their specific dataset and it is not capable to model visual identity. Moreover, it is significantly slower than our method, using several seconds to compute the descriptors of an image.

In the end, Model-C outperforms other Convolutional Block configurations we tried and the previous works in terms of precision. Also, it is the closest one to the Human and Oracle baselines. Although our model has one of the best perplexity value, other architectures like Model-D and Model-C outperform it. The perplexity is computed using the probability of choosing $x^P$ over $x^N$ as the similar icon to $x^R$, that's why its value is highly dependent on the formula used to compute the probability $\mathbb P$ from a distance $\mathcal D$. Due to that, we trust more the values of the precision when choosing our model while we still consider the perplexity. 
\begin{table}[!htb]
	\centering
	\begin{tabular}{@{\extracolsep{4pt}}l cc cc@{}} 
		\hline
		& \multicolumn{2}{c}{\textbf{Precision ($\mathcal P$)}} & \multicolumn{2}{c}{\textbf{Perplexity ($\mathcal Q$)}} \\ \cline{2-3}\cline{4-5} 
		\textbf{Model} & \textbf{Raw} & \textbf{Majority} & \textbf{Raw} & \textbf{Majority}\\ 
		\hline
		Humans & 0.771 & 0.842 & - & - \\ 
		Oracle & 0.859 & 1  & - & -\\ 
		\hline
		Garces~\cite{GarcesSIG2014} & 0.609  & 0.627 & 1.578 & 1.591 \\
		VGG19~\cite{Simonyan14} & 0.639  & 0.654 & 1.558  & 1.571  \\
		\hline
		Model-A & 0.519 & 0.521 & 1.603 & 1.617\\
		Model-B & 0.508 & 0.507 & 1.608 & 1.622\\
		Model-C & 0.671 & 0.702 & 1.543 & 1.556\\
		Model-D & 0.667 & 0.699 & 1.515 & 1.527\\
		\hline
		Best model & 0.706 & 0.738 & 1.555 & 1.568\\
		\hline
	\end{tabular}
	\caption{Comparison of the precision and perplexity of different models and methods. We can observe how the chosen method outperforms the rest comparing the precision and it is the closest one to the human ratings. On the other hand, perplexity values are highly dependent on the formula used to obtain probabilities from distances, while precision only depends on turker's answers. Due to that, our decision on choosing the best model has been more influenced by the results on the precision.}
	\label{tab:modelsResults}
\end{table}


\section{Results and Applications}\label{sec:results}

\begin{figure*}[!htb]
	\centering
	\includegraphics[width=0.9\linewidth]{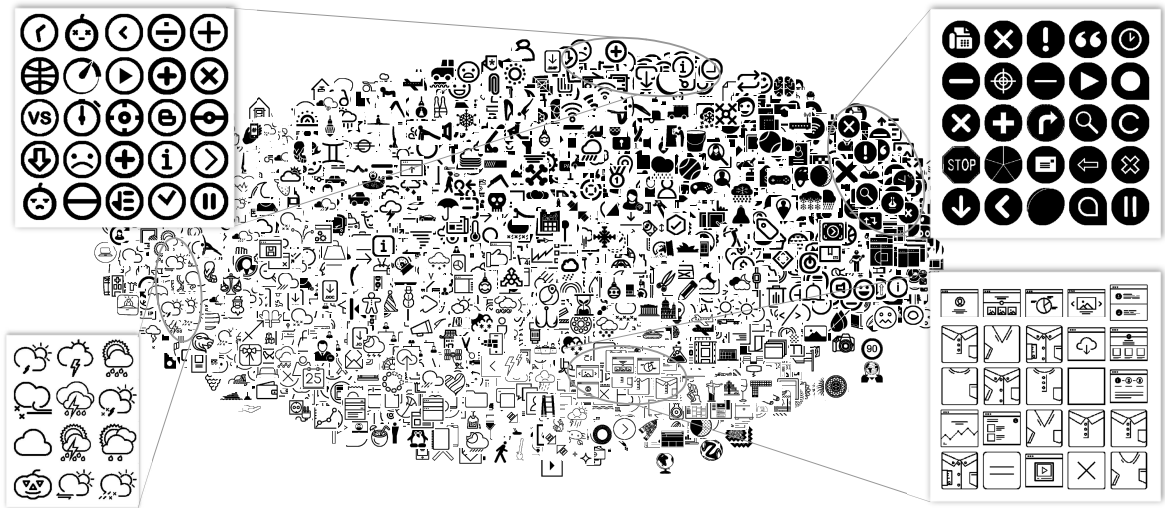}
	\caption{Visualization created using the t-SNE algorithm. It reduces the dimensionality of the feature vectors that our model learns to a two-dimensional Cartesian space. Note how icons with similar appearance are grouped in the same regions.}
	\label{fig:tsne}
\end{figure*}

The trained Siamese Neural Network is capable to produce high-quality embeddings in a new Euclidean feature space  which considers the properties of style and visual identity. 
We can visualize this space in 2D by using non-linear dimensionality reduction techniques, such as t-SNE~\cite{vanderMaaten08}. Results can be seen in Figure~\ref{fig:tsne}.

\paragraph{Comparison with Perceptual Kernels}

\begin{figure}[tb]
	\centering
	\includegraphics[width=\linewidth]{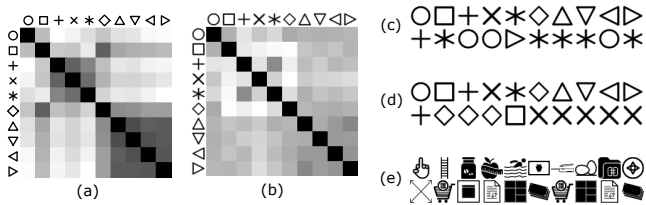}
	\caption{Comparison with the shape kernel of Demiralp et al.~\protect\cite{demiralp2014learning} (darker means more similar). \textit{(a)} Shape kernel of Demiralp et al. using ten gray-scale icons. \textit{(b)} Kernel obtained using our metric. Note that, as opposed to Demiralp's kernel, the triangles using our kernel are not invariant to rotation. In \textit{(c)} and (d) we show pairs of icons with maximum perceptual distances for Demiralp's kernel (c) and our metric (d). Our model is capable to return coherent icons with maximum perceptual distance although we did not collect the data with this specific purpose. On the other hand, the method of Demiralp et al. can only be computed for their set of ten icons.
		\textit{(e)} Pairs of icons with maximum distances using our whole dataset.}
	\label{fig:demiralpMAX}
\end{figure}

As we show in Equation~\ref{eq:distance}, for the same style, our metric measures the difference in visual identity, and, usually, this difference is linked to the shape of the object. Thus, we compare our metric with the perceptual kernel of Demiralp et al.~\cite{demiralp2014learning} which is optimized for shape similarity (Figure~\ref{fig:demiralpMAX}~\textit{(a)}). We take the same set of ten gray-scale icons, use our metric to compute the distances and normalize them between 0-1 range to obtain the matrix in Figure~\ref{fig:demiralpMAX}~\textit{(b)}. We also show in~\textit{(c)}, and~\textit{(d)} the icons with maximum distances with Demiralp's kernel and our distance $\mathcal{D}$, respectively. We observe that, although the results differ a little, both metrics perform very well in maximizing perceptual similarity. However, as opposed to Demiralp et al. work, our metric can be used with any input icon, while their kernel is strictly computed for that set of given icons. We additionally show in Figure~\ref{fig:demiralpMAX}~\textit{(e)} the icons with maximum distances in our whole dataset. Note that differences in style and visual identity are maximal.

\paragraph{Search by Similarity} 

\begin{figure*}[!htb]
	\parbox{.5\linewidth}{
		\centering
		\begin{tabular}{c c c c}
			\hline
			\rule{0pt}{3ex}    
			\textbf{Ref.} & \textbf{Our method} & \textbf{Garces et al.} & \textbf{VGG19}\\ \hline
			
			\rule{0pt}{12ex}  
			\raisebox{0.6\totalheight}{\includegraphics[width=0.1\linewidth]{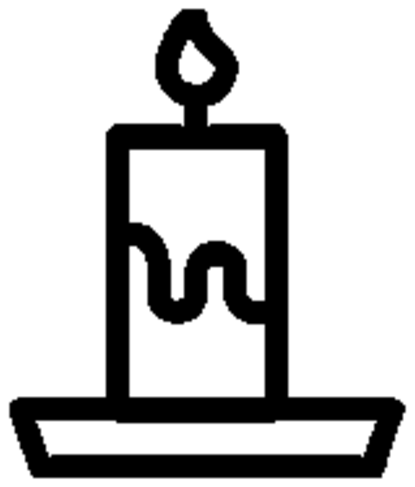}} & 
			\includegraphics[width=0.23\linewidth]{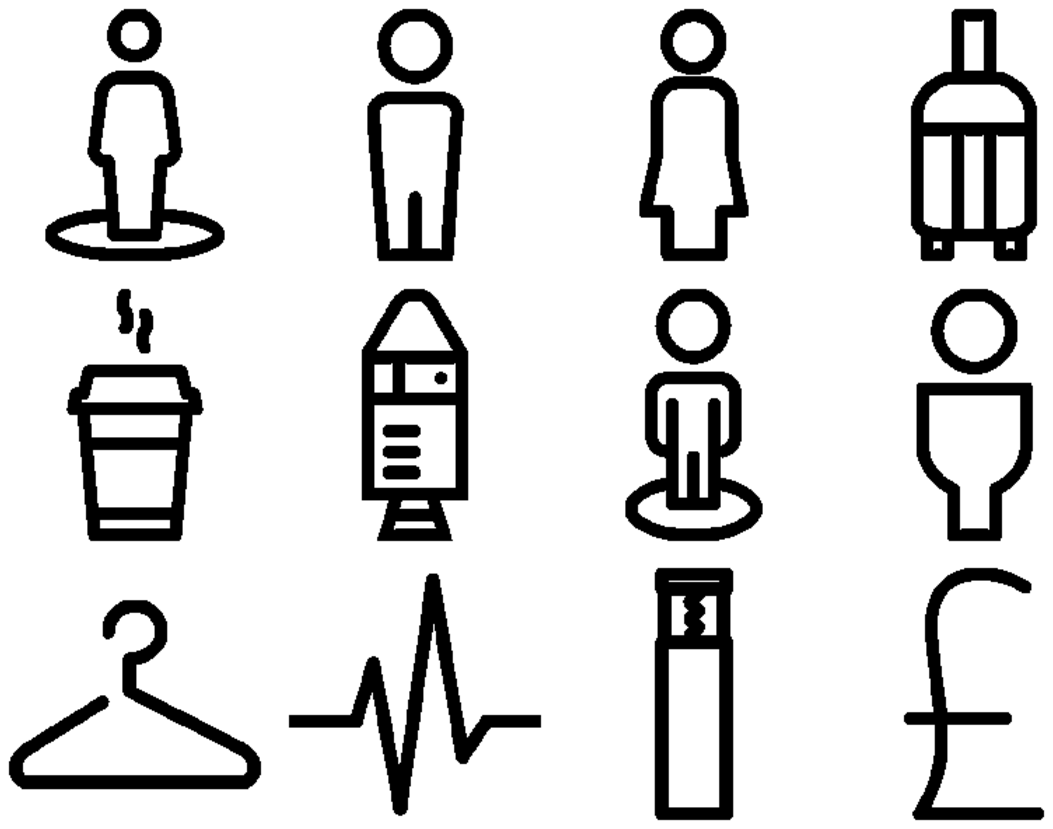} & 
			\includegraphics[width=0.23\linewidth]{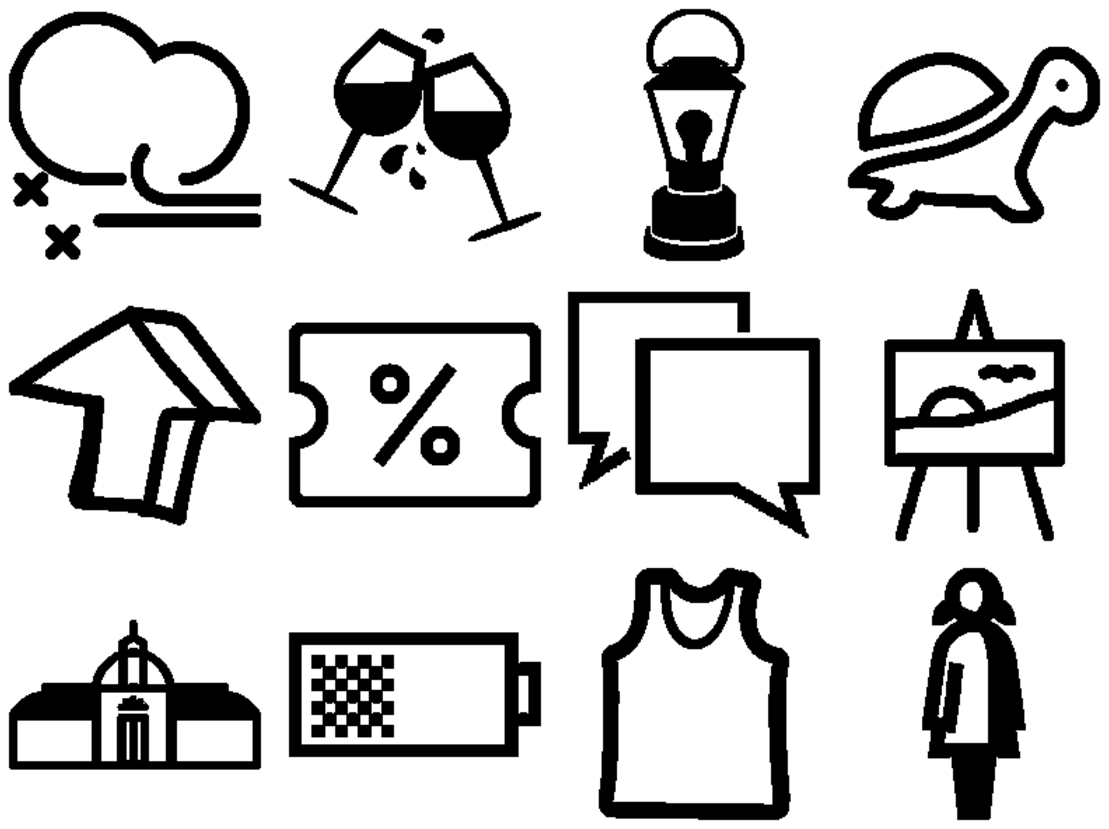}& 
			\includegraphics[width=0.23\linewidth]{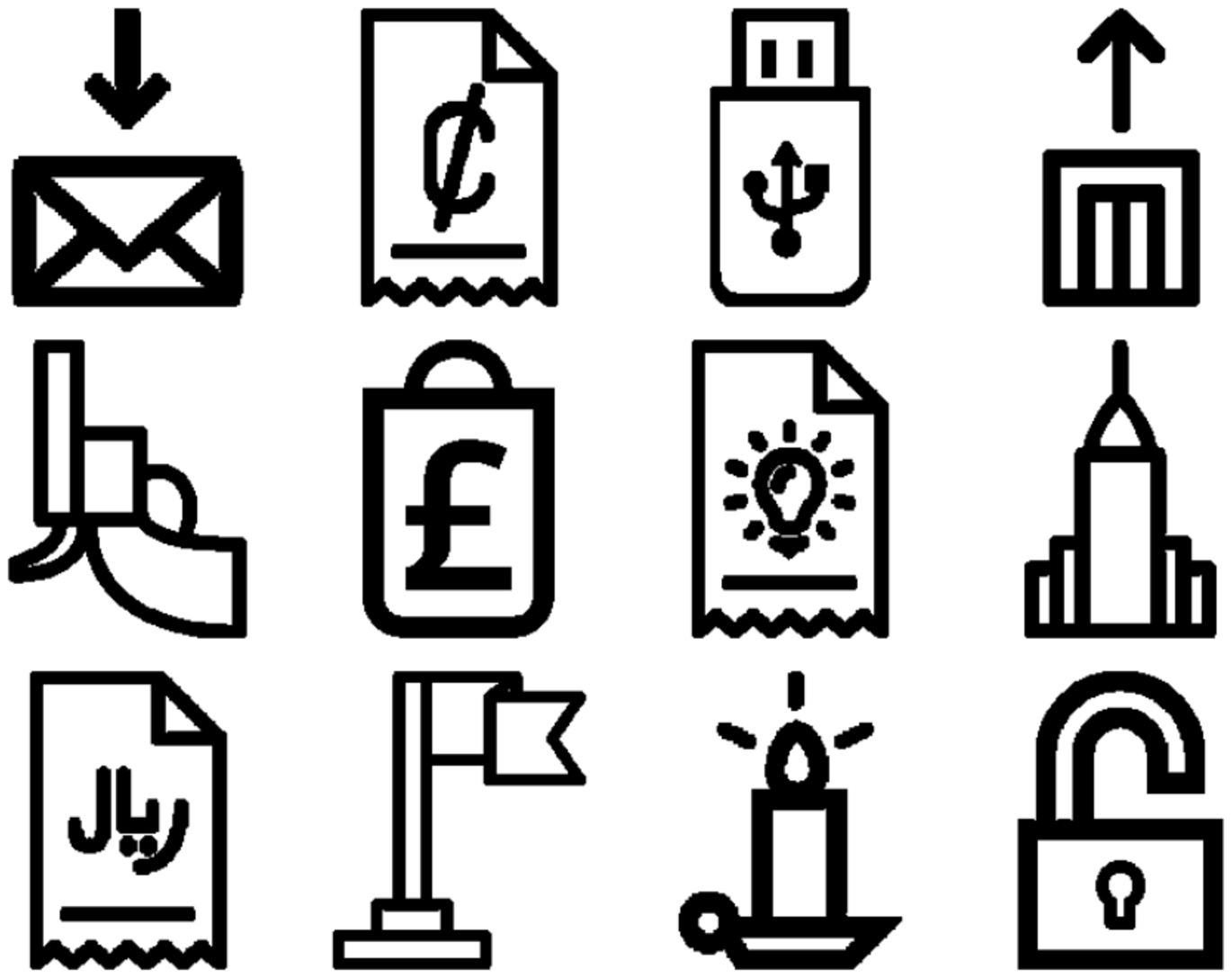}\\ \hline
			
			\rule{0pt}{12ex} 
			\raisebox{0.5\totalheight}{\includegraphics[width=0.1\linewidth]{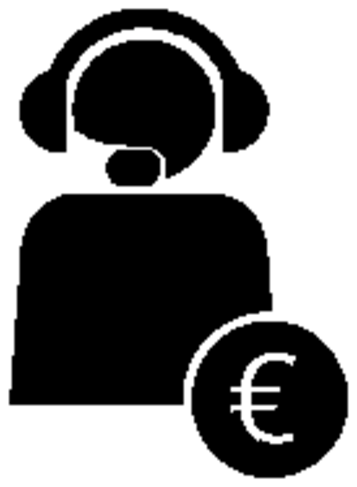}} & 
			\includegraphics[width=0.23\linewidth]{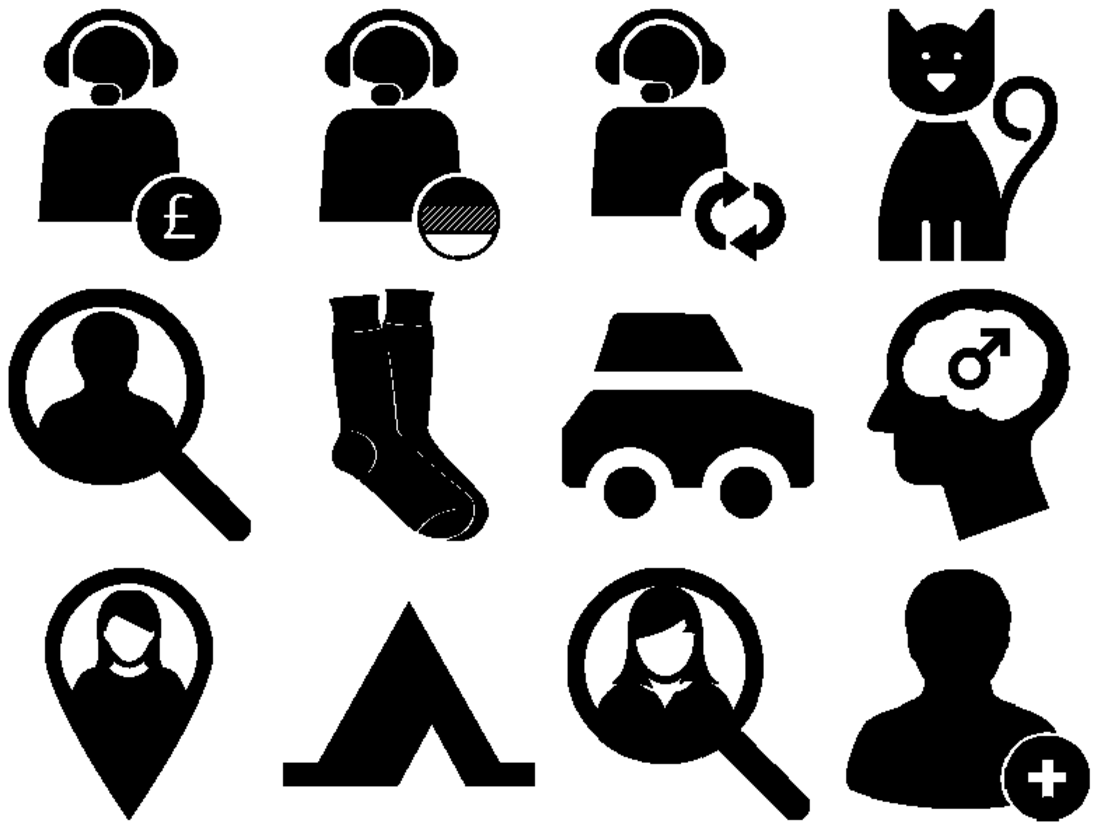} & 
			\includegraphics[width=0.23\linewidth]{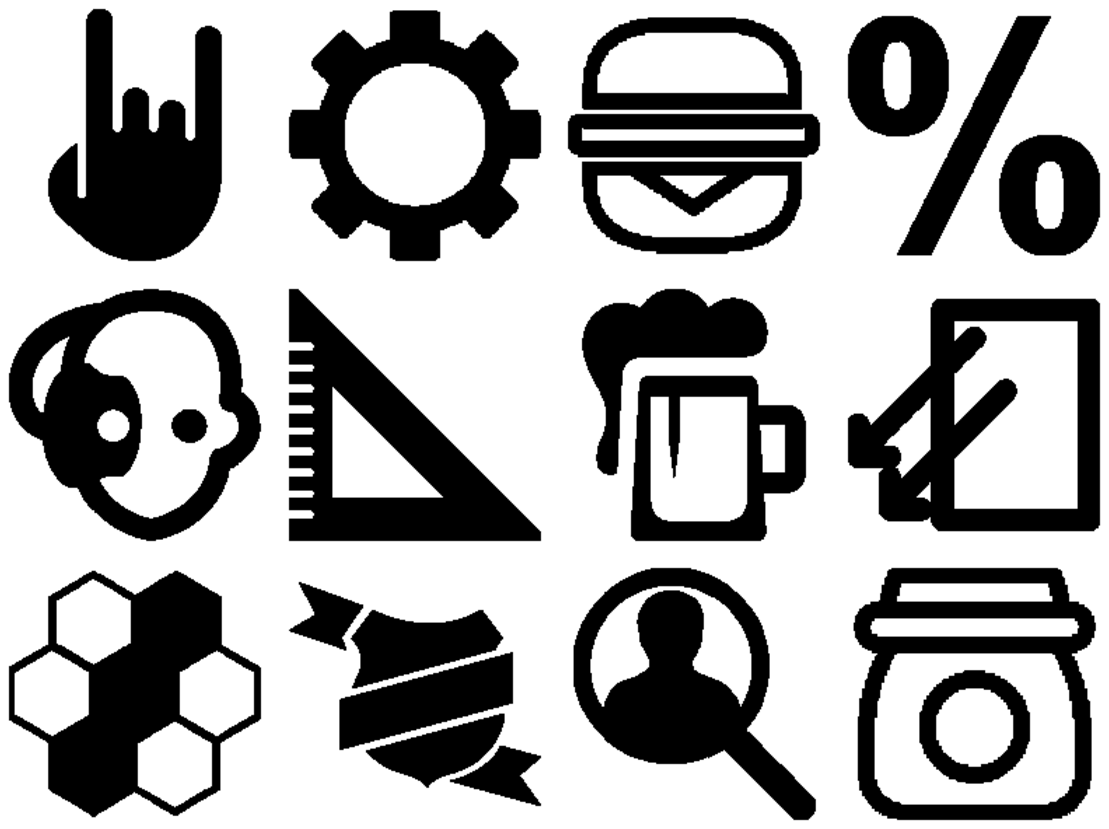}& 
			\includegraphics[width=0.23\linewidth]{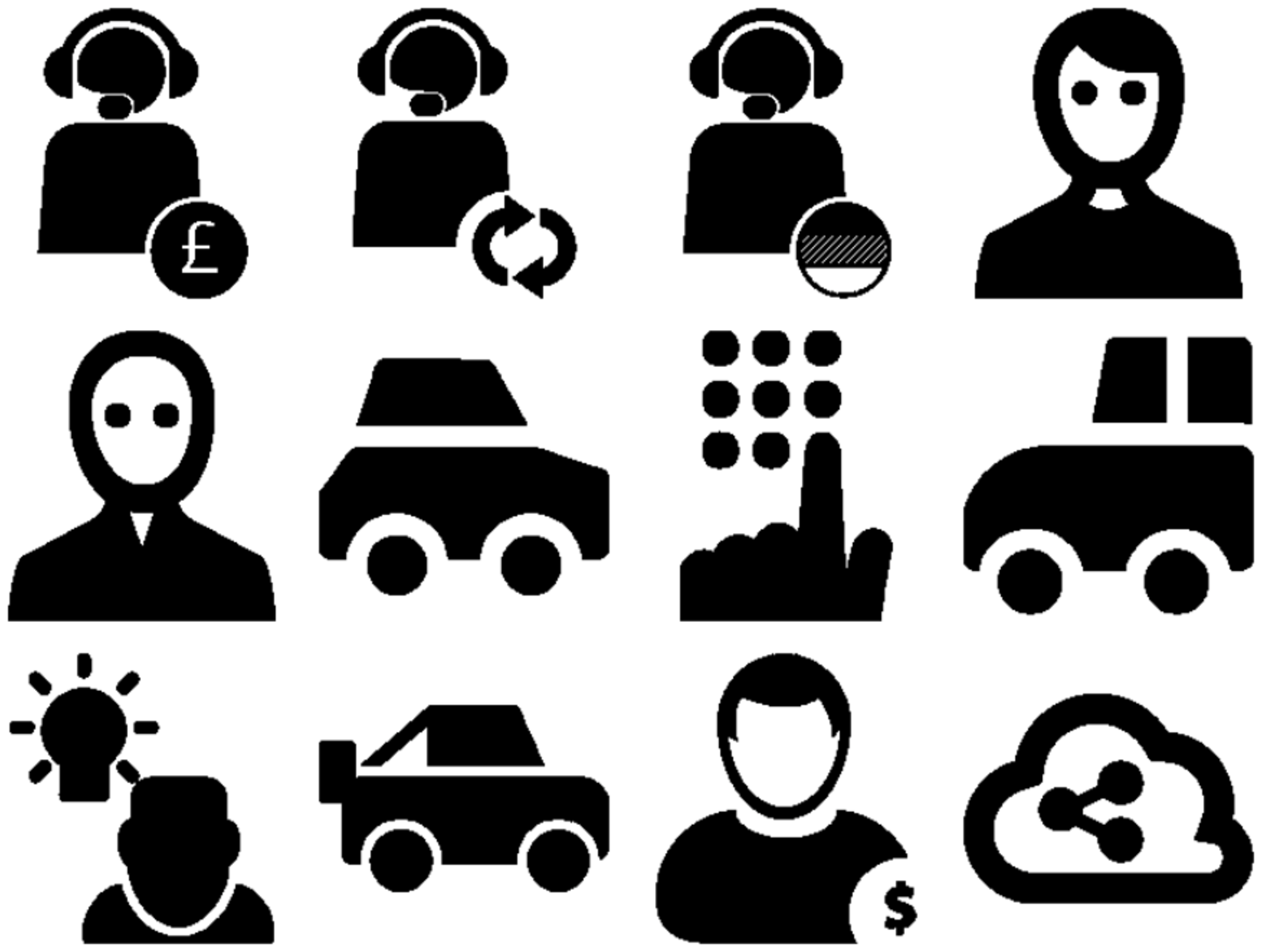}\\ \hline
			
			\rule{0pt}{12ex}
			\raisebox{0.5\totalheight}{\includegraphics[width=0.1\linewidth]{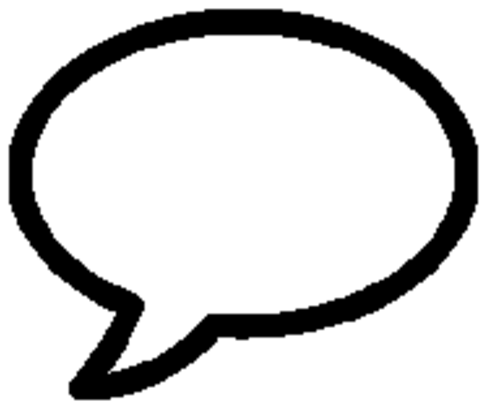}} &
			\includegraphics[width=0.23\linewidth]{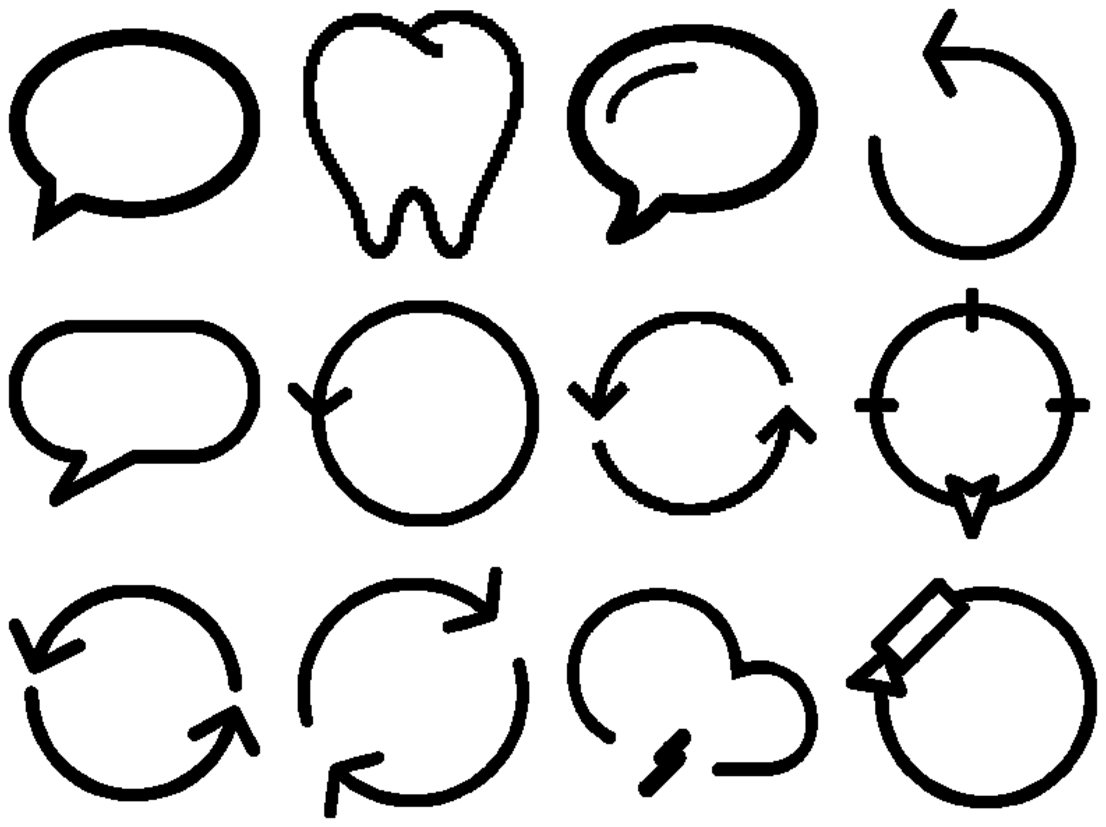} & 
			\includegraphics[width=0.23\linewidth]{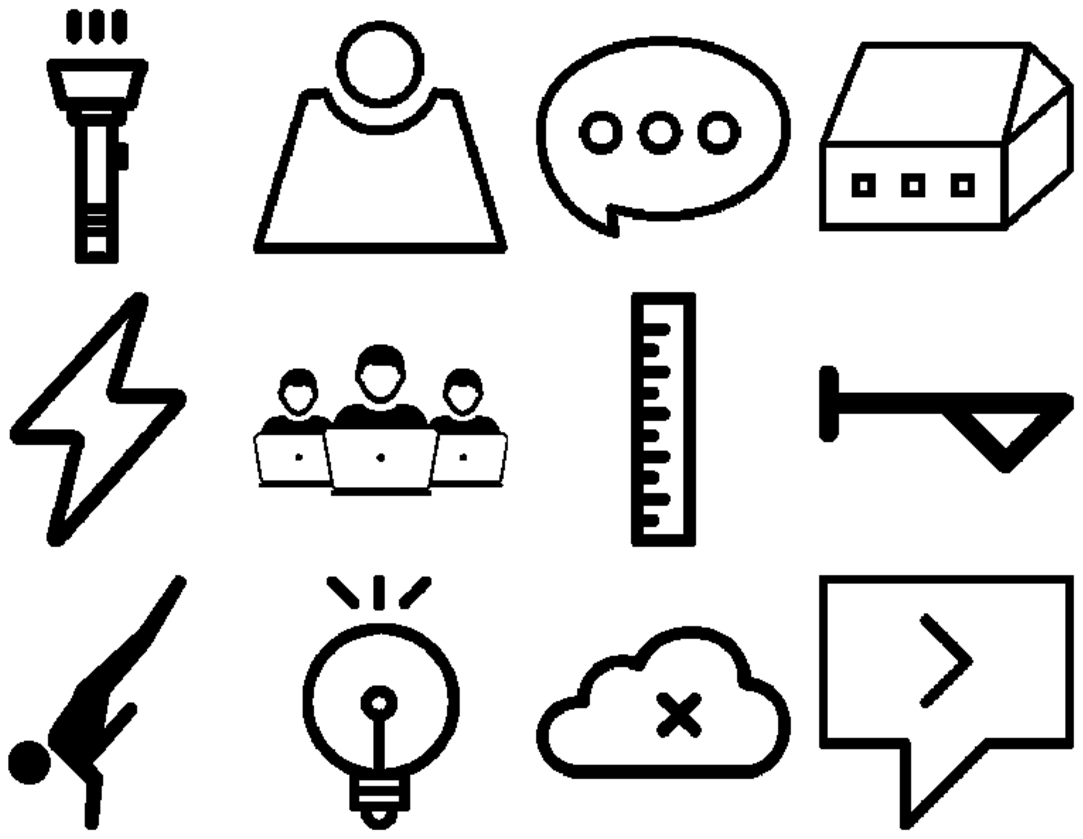}& 
			\includegraphics[width=0.23\linewidth]{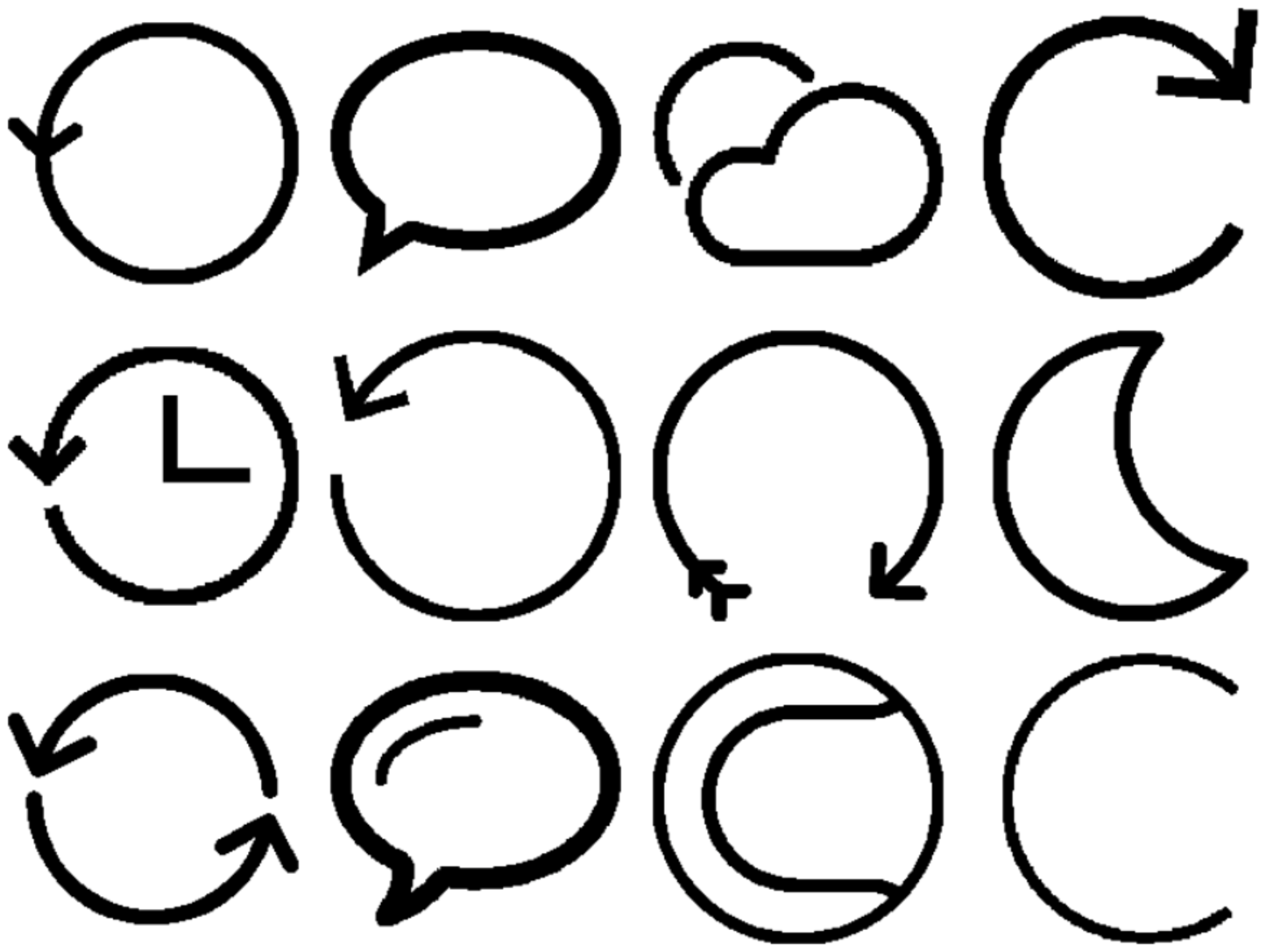}\\ \hline
		\end{tabular}
	}
	\hfill
	\parbox{.5\linewidth}{
		\centering
		\label{tab:resultsComparison2}
		\begin{tabular}{c c c c}
			\hline
			\rule{0pt}{3ex}    
			\textbf{Ref.} & \textbf{Our method} & \textbf{Garces et al.} & \textbf{VGG19}\\ \hline
			
			\rule{0pt}{12ex} 
			\raisebox{0.5\totalheight}{\includegraphics[width=0.1\linewidth]{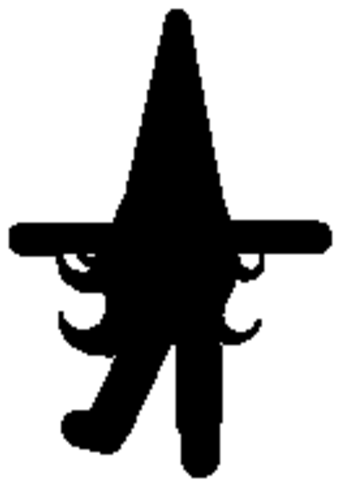}} & 
			\includegraphics[width=0.23\linewidth]{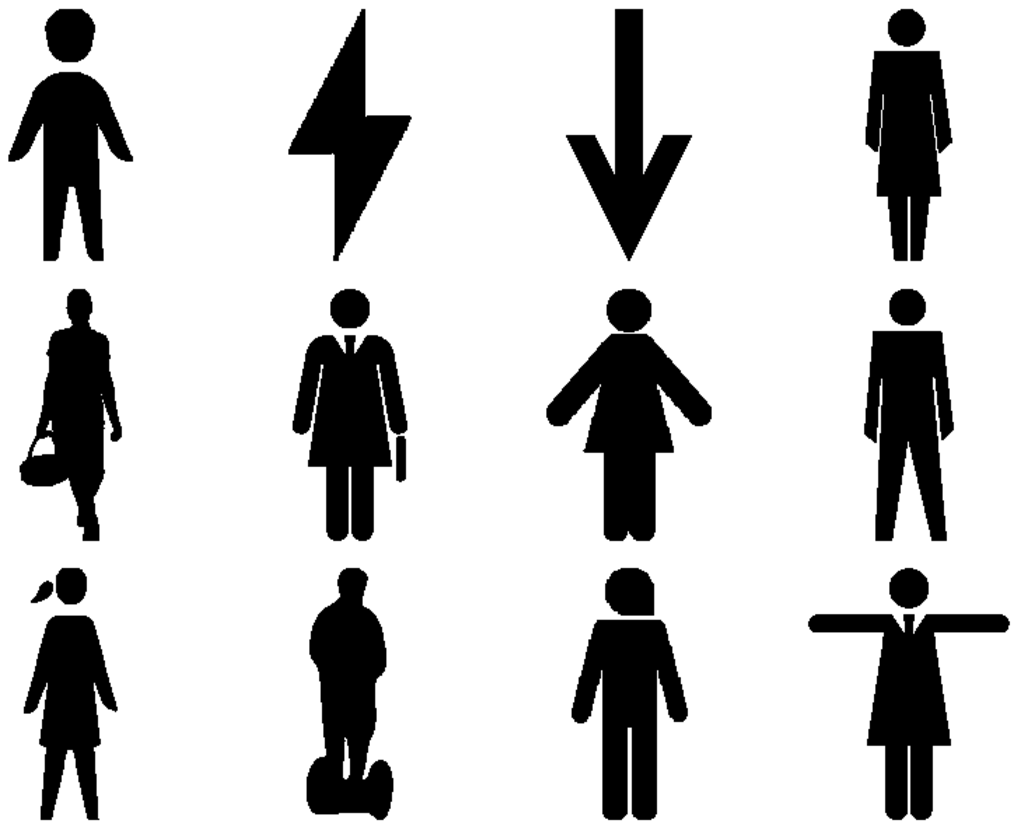} & 
			\includegraphics[width=0.23\linewidth]{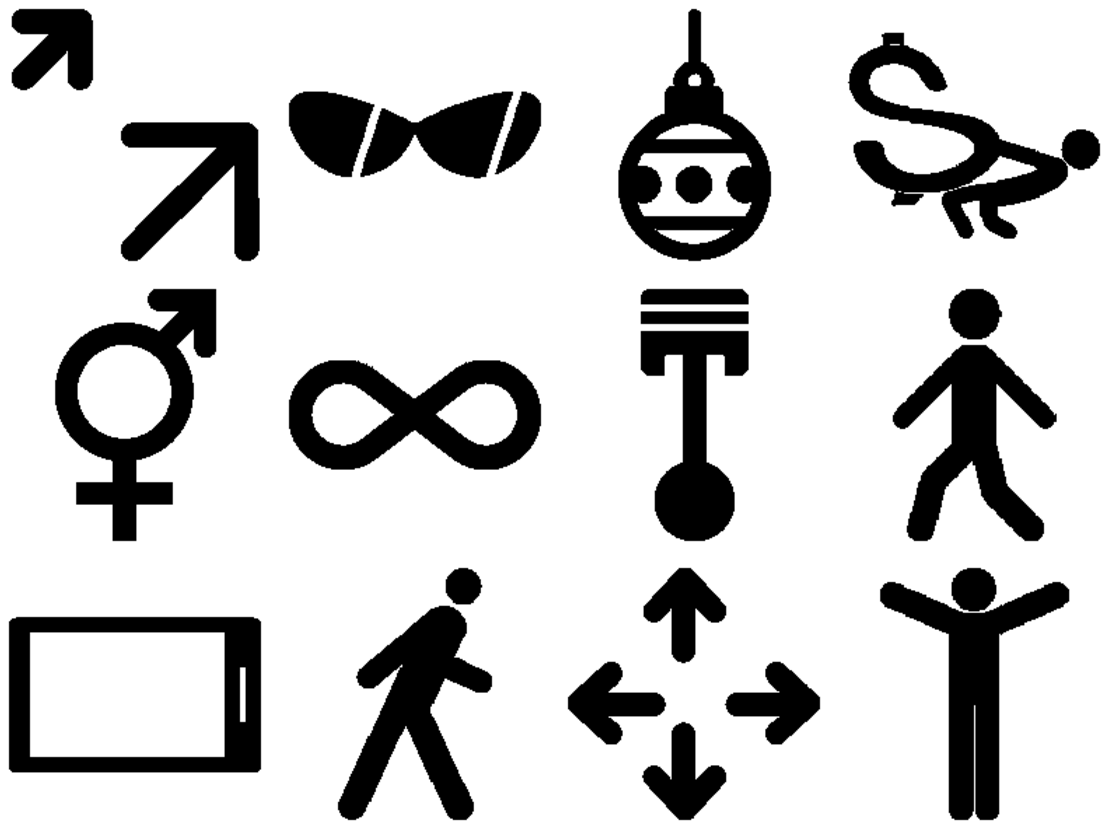}& 
			\includegraphics[width=0.23\linewidth]{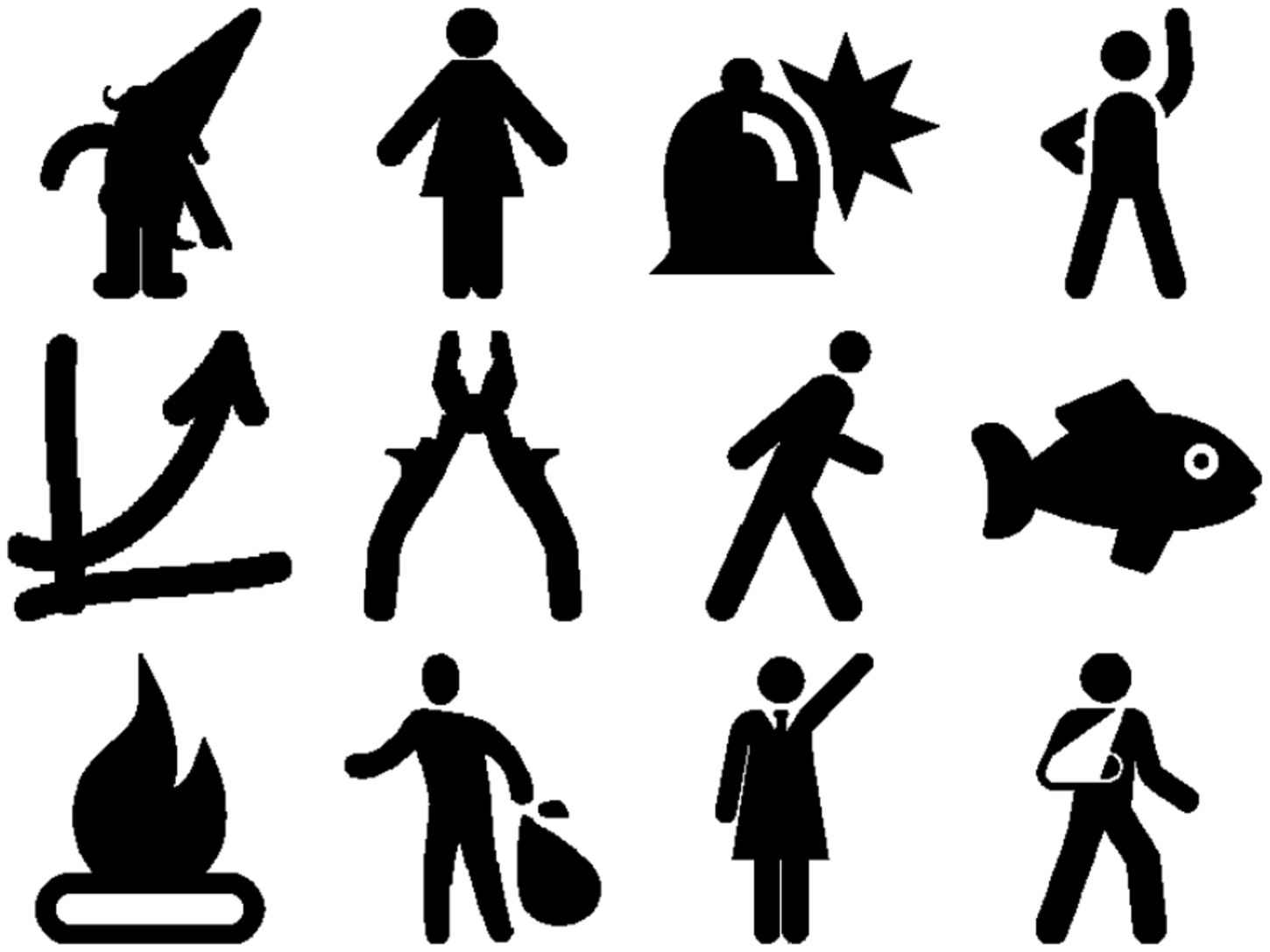}\\ \hline
			
			\rule{0pt}{12ex}  
			\raisebox{0.5\totalheight}{\includegraphics[width=0.1\linewidth]{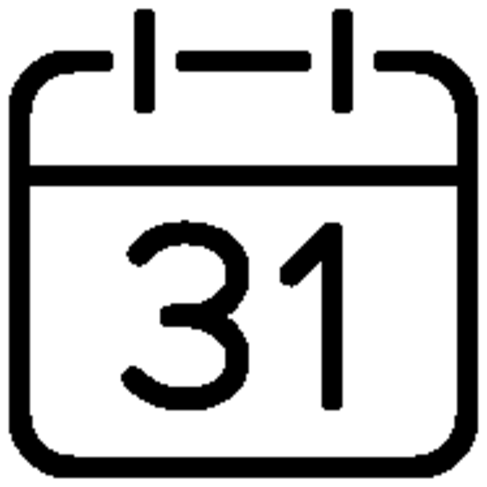}} & 
			\includegraphics[width=0.23\linewidth]{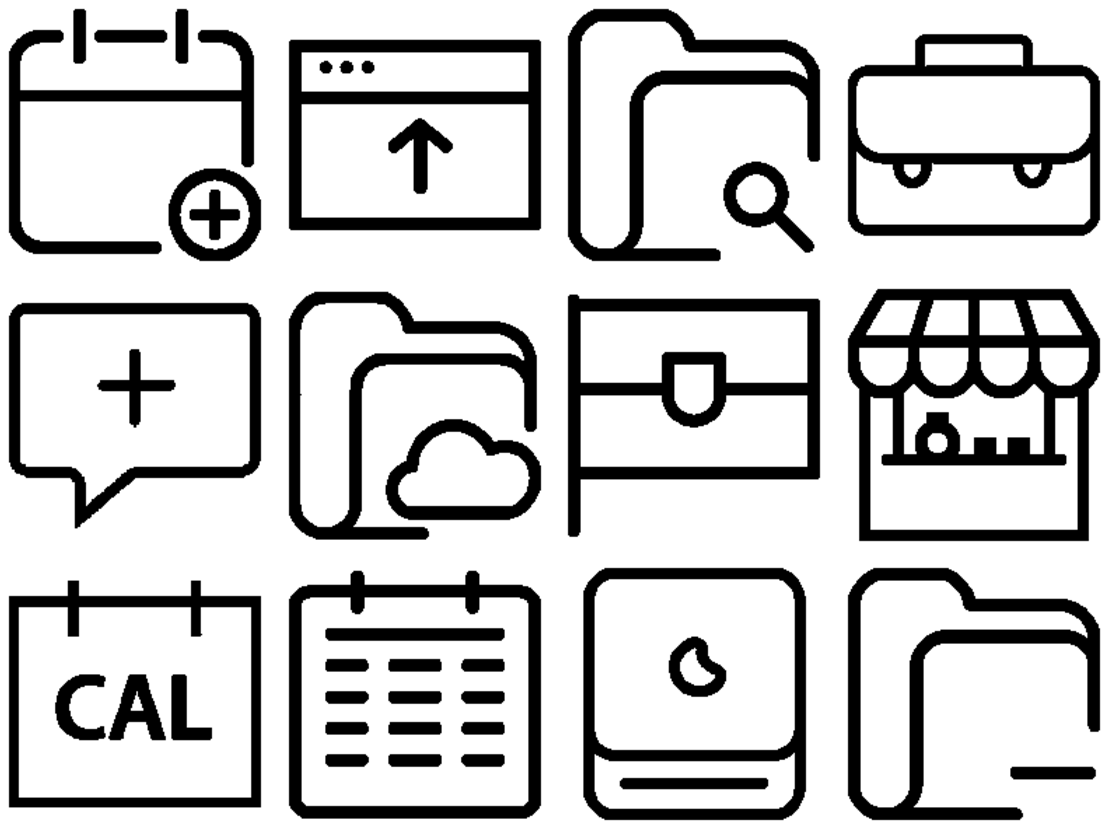} & 
			\includegraphics[width=0.23\linewidth]{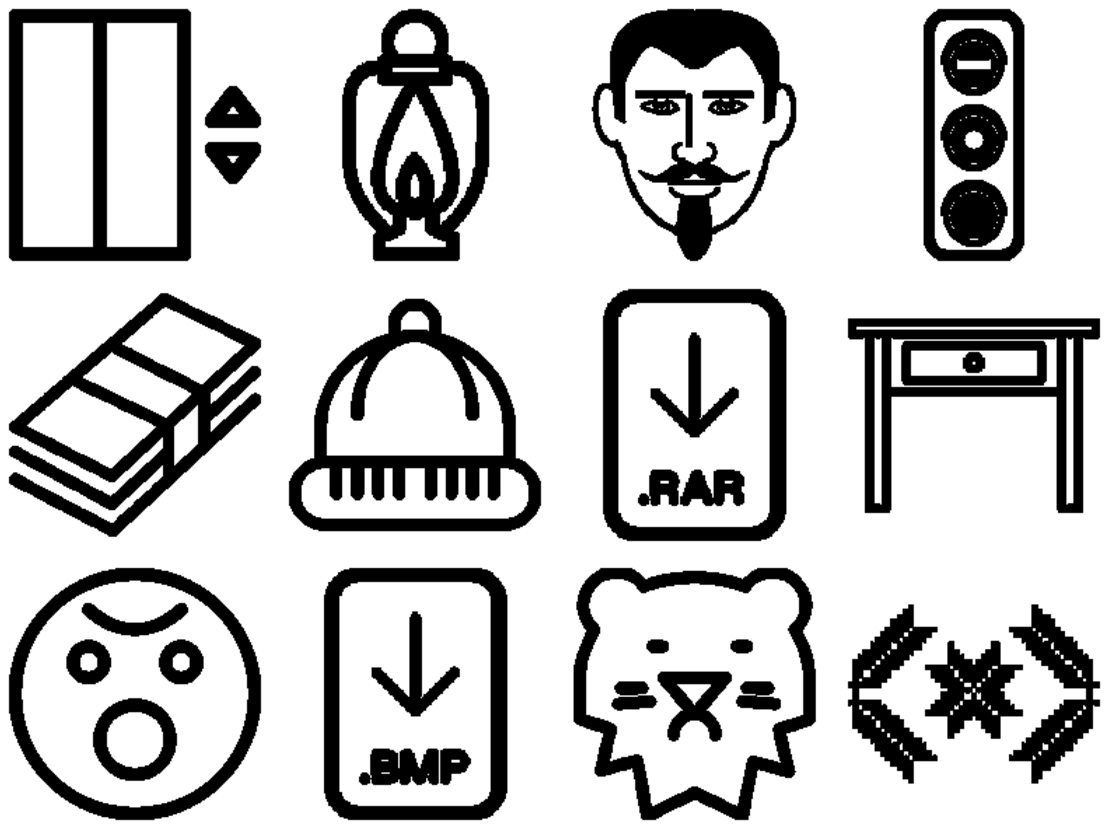}& 
			\includegraphics[width=0.23\linewidth]{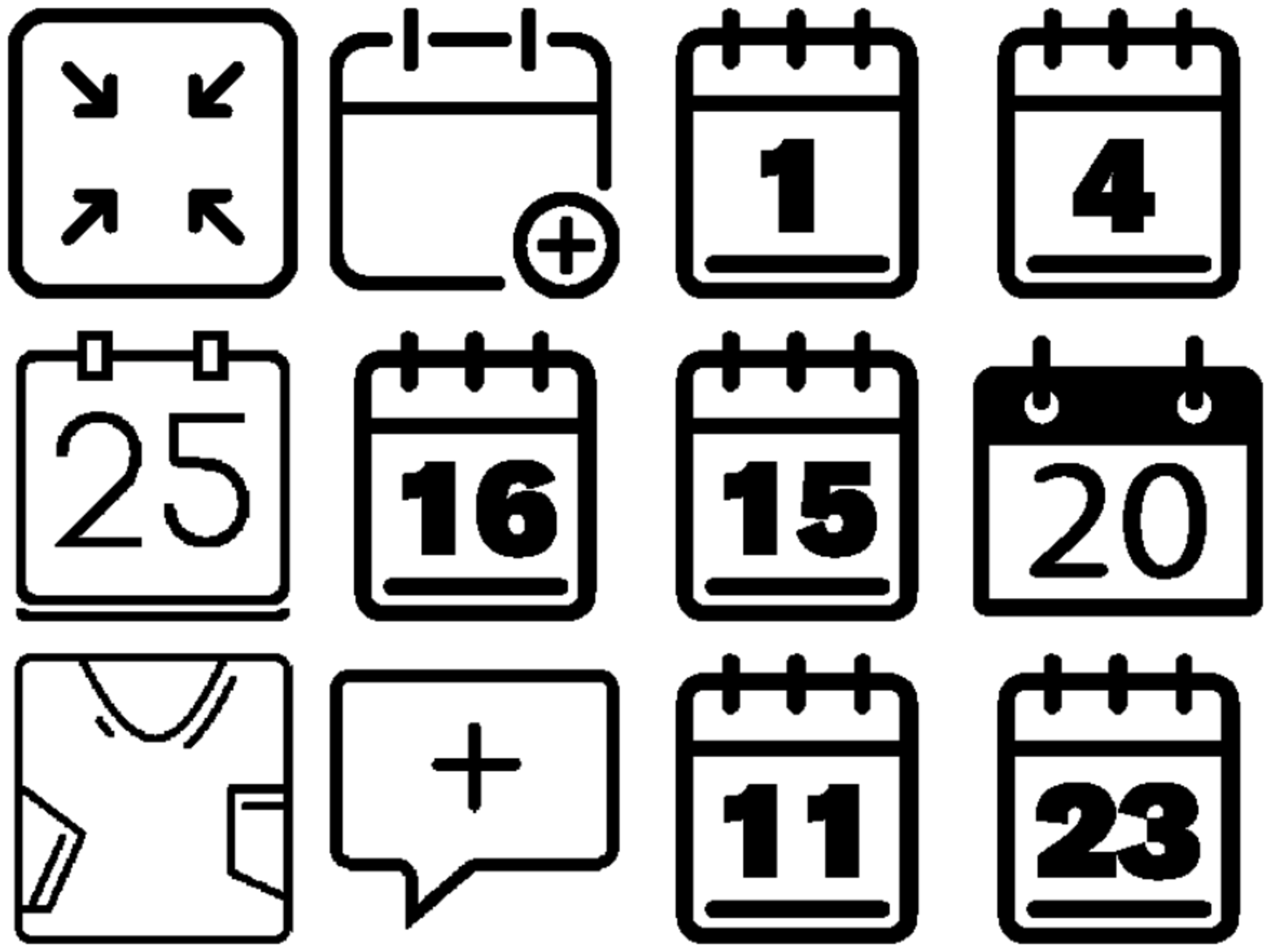}\\ \hline
			
			\rule{0pt}{12ex} 
			\raisebox{0.5\totalheight}{\includegraphics[width=0.1\linewidth]{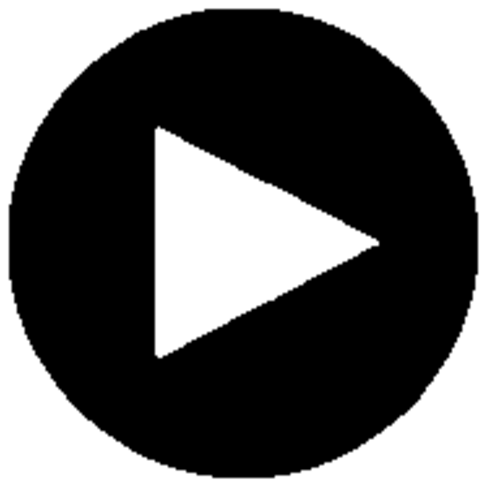}} & 
			\includegraphics[width=0.23\linewidth]{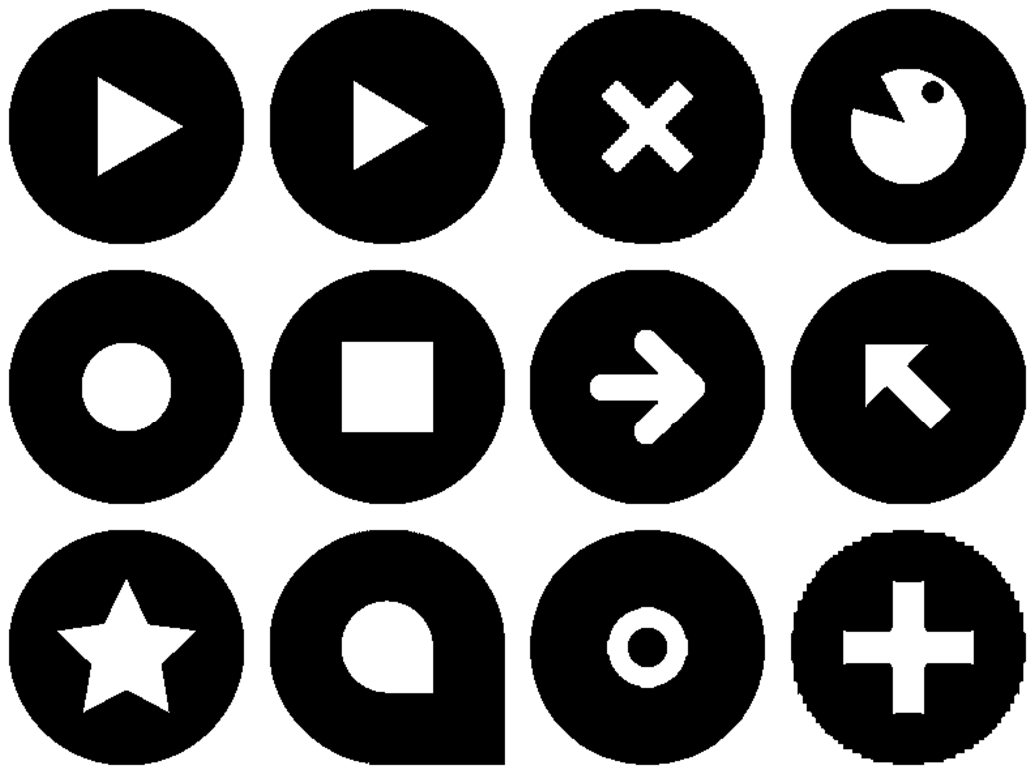} & 
			\includegraphics[width=0.23\linewidth]{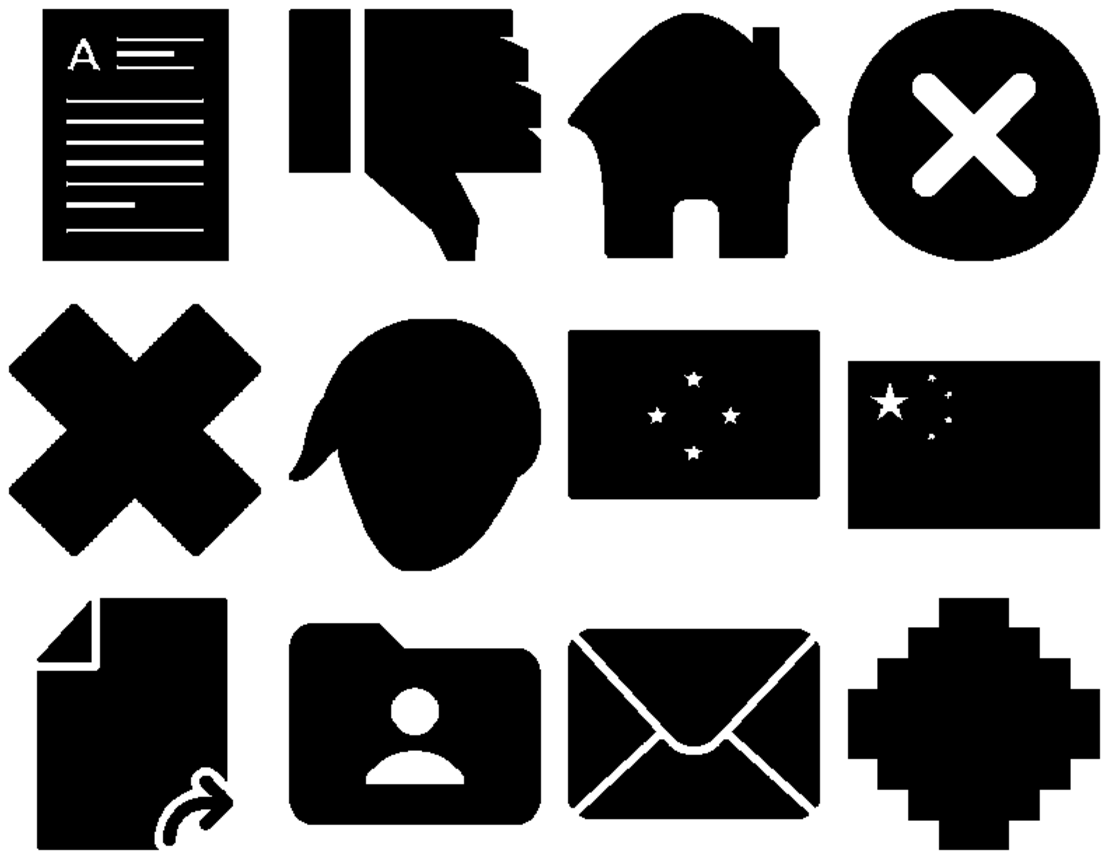}& 
			\includegraphics[width=0.23\linewidth]{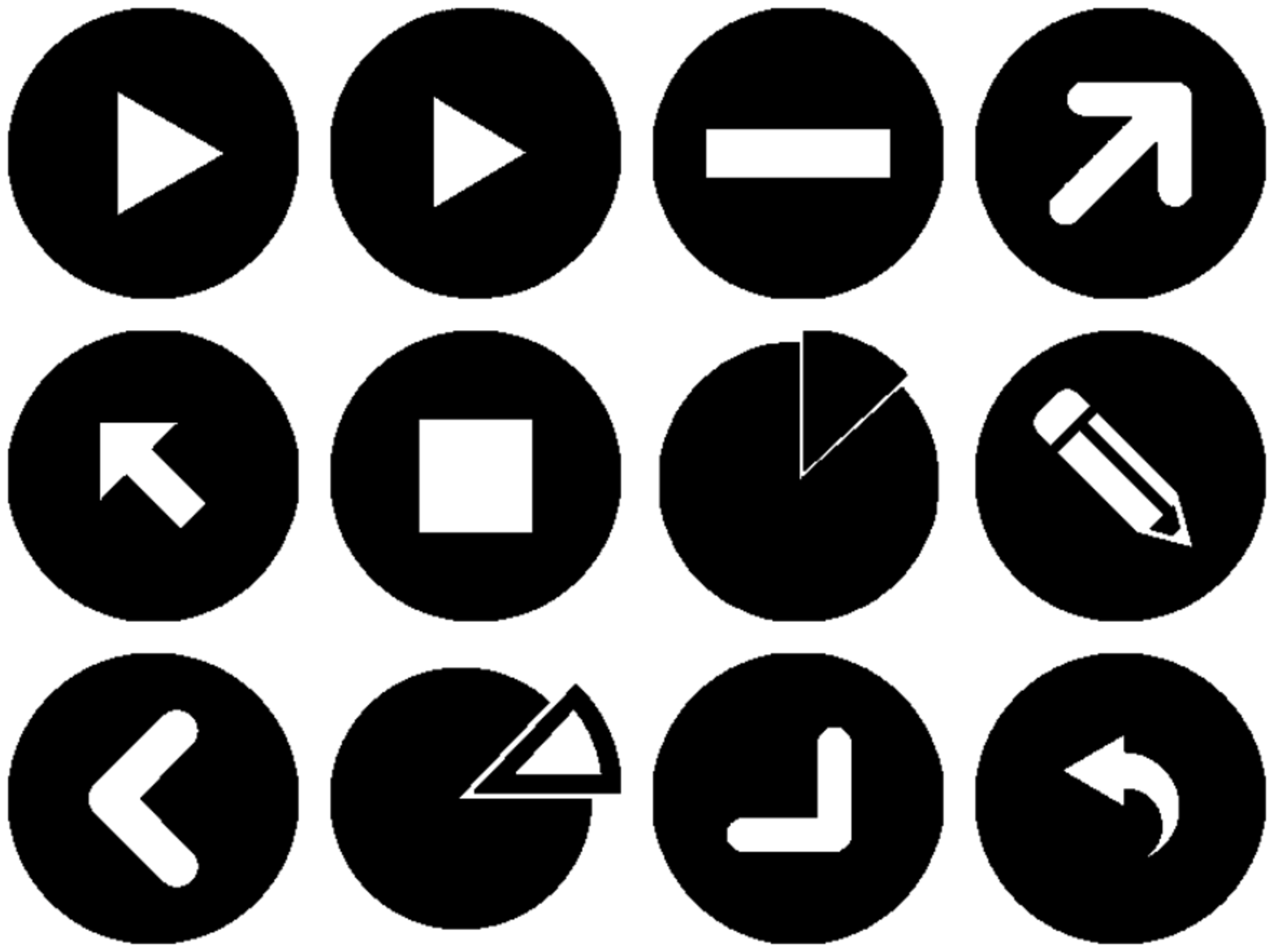}\\ \hline
	
		\end{tabular}
	}
	\caption{The following figure shows the most similar images given a reference and compares the output given by our method with the output given by Garces et al.~\cite{GarcesSIG2014} and the pretrained network VGG19~\cite{Simonyan14}. We can observe how our method returns visually appealing results considering both style and visual identity. The method of Garces et al. returns icons that match the style of the reference in most cases yet it does not consider visual identity. Some of the results obtained with the network VGG19 are coherent in style and visual identity (circles), however, several icons do not match the style of the reference (candle, calendars). Moreover, the network VGG19 encodes each input icon in a 4096-dimensional space and uses 144M parameters while our method encodes each icon into a 256-dimensional space and uses 47M parameters.}
	\label{fig:resultsComparison}
\end{figure*}

Our distance metric allows search by similarity. Given a query icon, we can search the k-nearest neighbors over the entire icon dataset.
Results are shown in Figure~\ref{fig:resultsComparison}. We compare our results with the output given by the method presented by Garces et al.~\cite{GarcesSIG2014} and the pretrained network VGG19~\cite{Simonyan14}. We can notice that while Garces et al. performs reasonably well to capture low-level style features like strokes and fills, it fails at higher-level elements, and the visual identity is not captured. This is due to the fact that their hand-crafted feature space does not include any feature to capture shape.
The network VGG19 after being trained with millions of images can be used as a powerful image descriptor thanks to the knowledge it acquired regarding image features like contours, textures or shapes. The results of VGG19 seem to have coherent visual identity yet some fail in terms of style (see Figure~\ref{fig:resultsComparison} \textit{candle} and \textit{calendar} rows). This imprecision is also observable in the numerical evaluation of Section~\ref{sec:compPW}. 

\paragraph{Optimized Icon Sets}

\begin{figure}[!htb]
	\centering
	\includegraphics[width=0.90\linewidth]{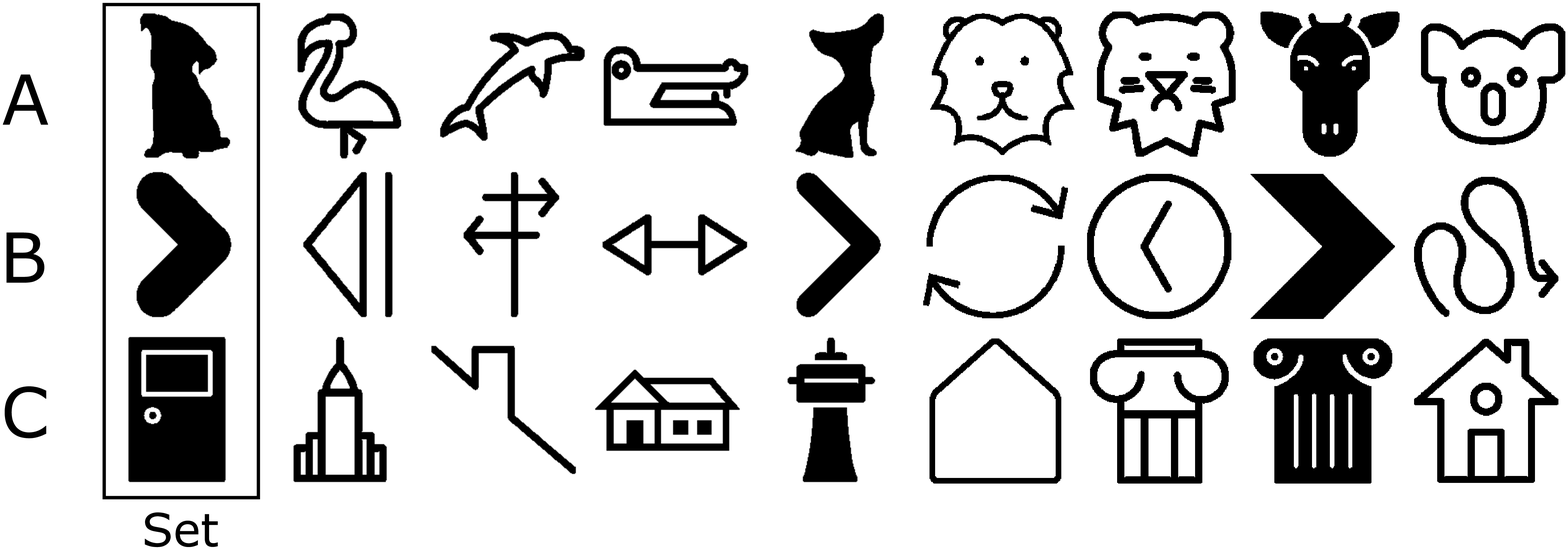}
	\caption{General icon set proposal for the keywords: animals (A), arrows (B) and buildings (C). Sets are optimized for the properties of visual identity and style using our method.}
	\label{fig:tripletsCandidate}
\end{figure}

Our method can be useful helping designers in creating applications or graphical user interfaces. Given a set of semantic keywords, we can propose icon sets optimized for the properties of style and visual identity.
In the example of Figure~\ref{fig:tripletsCandidate}, we choose the keywords \textit{animals (A)}, \textit{arrows (B)} and \textit{buildings (C)} and we obtain three sets of icons $\{x_A\}$, $\{x_B\}$, $\{x_C\}$ with 36, 112, and 55 elements, respectively. 
We define a candidate icon set as a triplet $(x_A, x_B, x_C) \in T$, where $T$ is the set containing all the possible combinations of icons for the selected keywords (note that we decided to have triplets as icon sets, but this could arbitrarily grow to icon sets of $n$ elements with $n\in[1, \infty]$).
For this case, $T$ contains 
more than $2\cdot10^6$ possible triplets. 
The goal is to find: 
$argmin_{i,j,k}\;\mathcal D_{set}(x_{A_i}, x_{B_j}, x_{C_k})$,
where $\mathcal{D}_{set} (x_A, x_B, x_C) = \mathcal{D}(x_A, x_B) + \mathcal{D}(x_B, x_C) + \mathcal{D}(x_A, x_C)$. 
The candidate sets are those whose distances are minimal. As we can see in the figure, the proposed icon sets are highly coherent.

\begin{figure}[!htb]
	\centering
	\includegraphics[width=\linewidth]{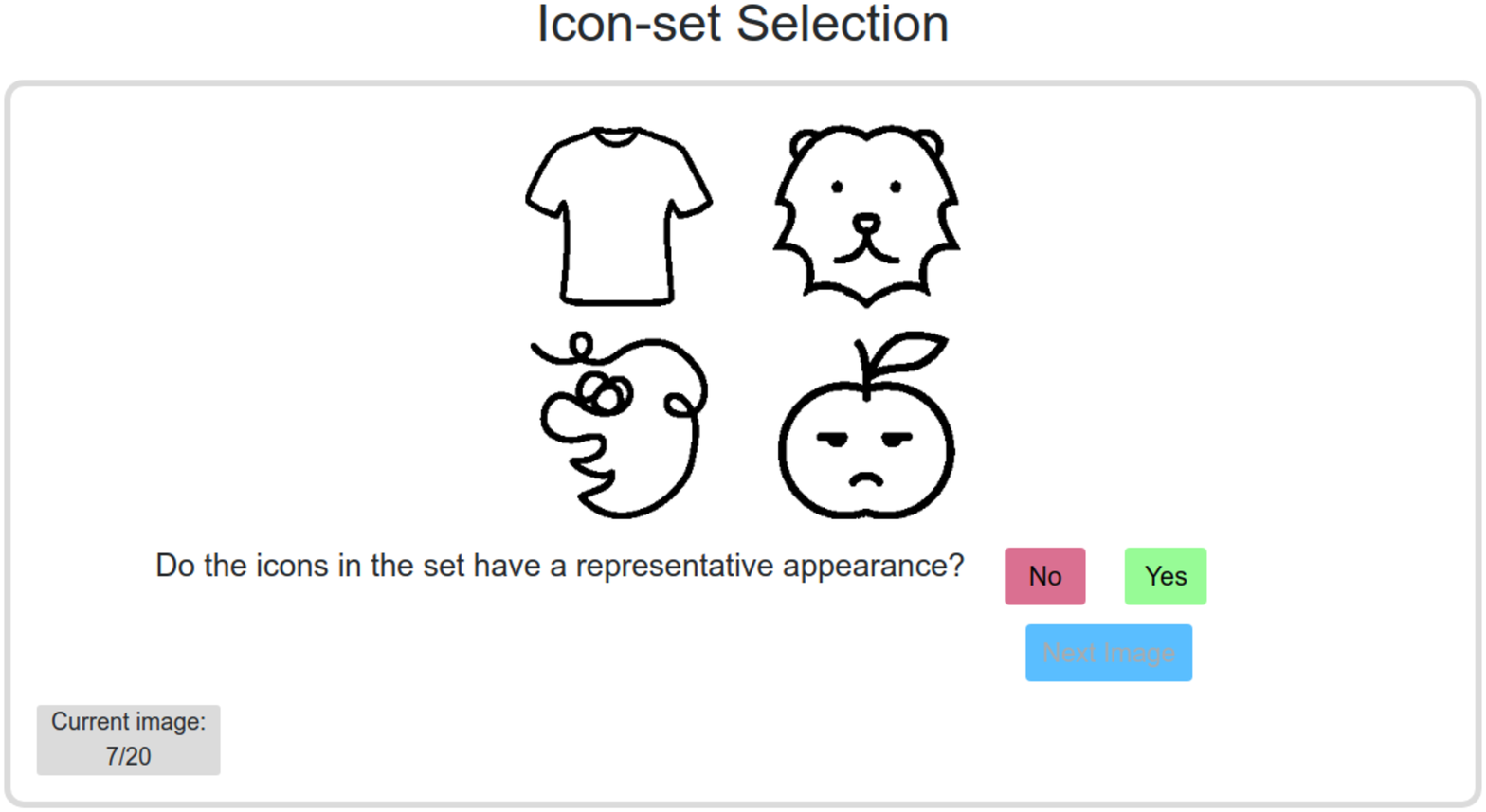}
	\caption{Screenshot of the test developed to validate the usefulness of the proposed icon sets. The icon set is made of four icons belonging to the keywords: clothes (top-left), animal (top-right), faces (bottom-left) and food (bottom-right). Below the images the question appears allowing for a binary answer (yes or no). The blue button goes to the next icon set and on the bottom left corner, whit gray background, we can see the progress of the test.}
	\label{fig:setsValidation}
\end{figure}

To evaluate how useful the proposed \textit{optimized icon sets} are, we gather subjective judgments from annotation experts. We show several optimized icon sets to the rater and ask her two questions: "\textit{Do the icons in the set have a representative appearance?}". The human-rater can only answer either yes or no. We created 100 sets using the method previously explained and 20 randomly sampling icons. Each survey contains 20 icon sets to be evaluated, 16 randomly sampled from the set of 100 created with our method and 4 randomly sampled from the set of random icon sets. Each icon set is made by four icons belonging to four different keywords. The keywords are also randomly sampled from a group of 9 candidates (animals, arrows, buildings, clothes, food, faces, music, humans and documents). Each keyword contains around 80 different icons from the test set with a wide variety of styles and visual identities. The Figure~\ref{fig:setsValidation} shows a screenshot of the test carried out to validate the proposed icon sets. At the end we collected 25 subjective evaluations from raters with previous experience in Computer Graphics or Graphic Design, 8 raters are females and ages range between 20 to 32 years old with an average of 25 years old. Raters thought the visual appearance of the icons is representative within the sets returned by our method 75.25\% of the times. On the other hand, raters found the appearance of the set representative only 28\% of the times for sets with randomly sampled icons.

\section{Conclusion and Future Work}\label{chap:conclusion}

In this work, we have presented a model for measuring the properties of style and visual identity in iconography. 
As opposed to previous works, which only focus on low-level style features, our method is able to model high-level properties of the icons, capturing its visual identity.   
Our learned model maps each icon into a 256-dimensional feature space which allows direct comparisons by computing Euclidean distances.
We have shown that our metric can be used to ease 
the process of icon set selection for users. Moreover, our approach is generalizable and can be used with any image outside the initial dataset.

There are many avenues for research following our work. The most immediate extension is to take into
account color compatibility measures~\cite{odonovan2011} to automatically colorize the icons to a particular color style. Similarity metrics can also be used as a guide to evaluate content generation methods, in our case, our metric could be used in combination with the work of Liu et al.~\cite{Liu2016} to automatically iconify pictures according to a desired style. In this regard, 
the success of deep generative methods for style transfer in fonts~\cite{upchurchSB16} suggests that such kind of techniques could be applied in this domain too. Moreover,
Our network could be used in combination with semantic object labeling or object sketches to train better models that take into account object semantics besides depiction.

On the other hand, while CNNs have received a lot of attention for natural images, they are still highly unexplored for graphic designs. Since it is a domain with a simpler underlying representation, in theory, it should require less training data. 
We also believe that our work can inspire future works in the problem of extracting shape descriptors for 2D images. It is well known that Convolutional Neural Networks capture coarse shapes in the deeper layers of the hierarchy~\cite{zeiler2014visualizing}, but it is ongoing work to really understand how to disentangle this information to be used as a standalone shape descriptor.

\section*{Acknowledgements}

Acknowledgements:

We want to thank the anonymous reviewers and Adrian Jarabo for their insightful comments on the manuscript. This project has received funding from the European Research Council (ERC) under the European Union’s Horizon 2020 research and innovation programme (CHAMELEON project, grant agreement No 682080).


\bibliographystyle{acm}
\bibliography{bibliography}

\end{document}